\documentclass[sigconf, balance=false]{acmart}
\makeatletter
\def\balance{} 
\makeatother

\AtBeginDocument{%
  }

\usepackage{latexsym}

\usepackage{multirow}
\usepackage{makecell}
\usepackage{graphicx}   
\usepackage{multirow}
\usepackage{colortbl}
\usepackage{xcolor}

\usepackage{graphicx}
\usepackage{newfloat}
\usepackage{listings}

\usepackage{wrapfig}
\DeclareCaptionStyle{ruled}{labelfont=normalfont,labelsep=colon,strut=off} 
\floatstyle{ruled}
\newfloat{listing}{tb}{lst}{}
\floatname{listing}{Listing}
\usepackage{tabularx}   

\usepackage{pifont}     
\usepackage{textcomp}   

\usepackage{graphicx}
\usepackage{subcaption} 

\usepackage{multirow}
\usepackage{caption}
\usepackage{adjustbox}  
\usepackage{pdflscape}
\usepackage[utf8]{inputenc}  
\usepackage{comment}
 \usepackage{tikz}
\usetikzlibrary{patterns, calc, positioning, arrows.meta, shapes.geometric}
\usepackage{pgfplots}
\pgfplotsset{compat=1.18}
\usepackage{graphicx}

\usepackage{hyperref}
\usepackage{fontawesome5} 

\usepackage{adjustbox}

\usepackage[utf8]{inputenc}
\usepackage{siunitx}
\usepackage{tikz}
\usepackage{pgfplots}
\pgfplotsset{compat=1.18}
\usetikzlibrary{calc, positioning, arrows.meta, decorations.pathreplacing}
\usepackage{nameref}
\definecolor{c_nonlatin}{RGB}{230,159,0}
\definecolor{c_ocr}{RGB}{86,180,233}
\definecolor{c_math}{RGB}{204,121,167}
\definecolor{c_reason}{RGB}{0,114,178}
\definecolor{c_other}{RGB}{180,180,180}
\newcolumntype{C}{>{\centering\arraybackslash}X}
\usepackage{tcolorbox}
\usepackage{xcolor}
\usepackage{listings}
\usepackage{caption}
\tcbuselibrary{breakable, skins}
\definecolor{questionbg}{RGB}{245, 245, 250}
\definecolor{correctbg}{RGB}{230, 255, 230}
\definecolor{wrongbg}{RGB}{255, 235, 235}
\definecolor{analysisbg}{RGB}{255, 250, 240}

\tcbset{
    questionbox/.style={
        colback=questionbg,
        colframe=gray!70,
        boxrule=0.5pt,
        arc=2mm,
        left=2mm, right=2mm, top=2mm, bottom=2mm
    },
    responsebox/.style={
        colback=white,
        colframe=blue!50,
        boxrule=0.5pt,
        arc=2mm,
        fonttitle=\bfseries,
        title=#1
    },
    analysisbox/.style={
        colback=analysisbg,
        colframe=orange!70,
        boxrule=0.5pt,
        arc=2mm,
        fonttitle=\bfseries,
        title=Analysis
    }
}
\newcommand{\cmark}{\ding{51}}   
\newcommand{\xmark}{\ding{55}}   

\hypersetup{
    colorlinks=true,
    linkcolor=blue,
    filecolor=blue,      
    urlcolor=blue,
    citecolor=blue
}

\hypersetup{
    colorlinks=true,
    linkcolor=blue,
    filecolor=blue,      
    urlcolor=blue,
    citecolor=blue
}

\usepackage{tikz}
\usetikzlibrary{shapes.geometric, arrows.meta, positioning, calc, backgrounds, fit, decorations.pathreplacing}
\usepackage{pgfplots}
\usepackage{tikz}
\usepackage{colortbl}  

\pgfplotsset{compat=1.17}
\usepackage{hyperref}

\usepackage{hyperref}
\usepackage{fontawesome5}
\usepackage{xcolor}
\usepackage{hyperref}

\definecolor{linkgithub}{HTML}{24292e}
\newcommand{\hflink}[1]{\href{#1}{\faDatabaseSolid~\texttt{Dataset}}}
\definecolor{linkweb}{HTML}{4285F4}

\newcommand{\githublink}[1]{\href{#1}{\faGithub~\texttt{Code}}}
\usepackage{fontawesome5}
\usepackage{graphicx}
\usepackage{hyperref}

\newcommand{\resourcelinks}{%
  \vspace{0.5em}\par\noindent
  \centering
  \href{https://github.com/thisiskorea/EuraGovExam}{\faGithub~\texttt{Code}} \quad $\vert$ \quad
  \href{https://huggingface.co/datasets/EuraGovExam/EuraGovExam}{\raisebox{-0.1em}{\includegraphics[height=1em]{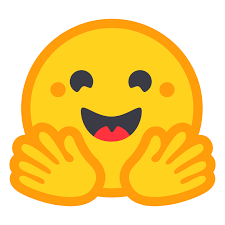}}~\texttt{Dataset}} \quad $\vert$ \quad
  \href{https://thisiskorea.github.io/EuraGovExam/}{\faGlobe~\texttt{Website}}%
}

\newcommand{\weblink}[1]{\href{#1}{\faGlobe~\texttt{Website}}}

\copyrightyear{2026}
\acmYear{2026}
\setcopyright{cc}
\setcctype{by}
\acmConference[KDD '26]{Proceedings of the 32nd ACM SIGKDD Conference on Knowledge Discovery and Data Mining V.2}{August 09--13, 2026}{Jeju Island, Republic of Korea}
\acmBooktitle{Proceedings of the 32nd ACM SIGKDD Conference on Knowledge Discovery and Data Mining V.2 (KDD '26), August 09--13, 2026, Jeju Island, Republic of Korea}
\acmDOI{10.1145/3770855.3817532}
\acmISBN{979-8-4007-2259-2/2026/08}
\begin{document}

\title{EuraGovExam: A Multilingual Multimodal Benchmark from Real-World Civil Service Exams}


\author{Jaeseong Kim}
\email{kjsqp1010@semyung.ac.kr}
\orcid{0009-0003-5625-259X}
\affiliation{%
  \department{School of Computer Science / Data Intelligence Lab}
  \institution{Semyung University}
  \city{Jecheon}
  \country{Republic of Korea}
}

\author{Chaehwan Lim}
\authornote{These authors contributed equally as co-second authors.}
\email{2024524103@semyung.ac.kr}
\orcid{0009-0006-0364-5930}
\affiliation{%
  \department{School of Computer Science / Data Intelligence Lab}
  \institution{Semyung University}
  \city{Jecheon}
  \country{Republic of Korea}
}

\author{Sang Hyun Gil}
\authornotemark[1]
\email{rlftkdgus75@semyung.ac.kr}
\orcid{0009-0006-6031-7503}
\affiliation{%
  \department{School of Computer Science / Data Intelligence Lab}
  \institution{Semyung University}
  \city{Jecheon}
  \country{Republic of Korea}
}

\author{Suan Lee}
\authornote{Corresponding author}
\email{suanlee@semyung.ac.kr}
\orcid{0000-0002-3047-1167}
\affiliation{%
  \department{School of Computer Science / Data Intelligence Lab}
  \institution{Semyung University}
  \city{Jecheon}
  \country{Republic of Korea}
}

\renewcommand{\shortauthors}{Kim et al.}
\begin{abstract}
We present \textbf{EuraGovExam}, a multilingual and multimodal benchmark sourced from real-world civil service examinations across five representative Eurasian regions: South Korea, Japan, Taiwan, India, and the European Union. Designed to reflect the authentic complexity of public-sector assessments, the dataset contains over 8,000 high-resolution scanned multiple-choice questions covering 17 diverse academic and administrative domains. Unlike existing benchmarks, EuraGovExam embeds all question content—including problem statements, answer choices, and visual elements—within a single image, providing only a minimal standardized instruction for answer formatting. This design demands that models perform layout-aware, cross-lingual reasoning directly from visual input. All items are drawn from real exam documents, preserving rich visual structures such as tables, multilingual typography, and form-like layouts. Evaluation results show that even state-of-the-art vision-language models (VLMs) achieve only 86\% accuracy, underscoring the benchmark's difficulty and its power to diagnose the limitations of current models. By emphasizing cultural realism, visual complexity, and linguistic diversity, EuraGovExam establishes a new standard for evaluating VLMs in high-stakes, multilingual, image-grounded settings. It also supports practical applications in e-governance, public-sector document analysis, and equitable exam preparation.

\resourcelinks
\end{abstract}



\maketitle

\begin{table*}[t]
\centering
\caption{Comparison of \textsc{EuraGovExam} with representative multimodal
and exam-based benchmarks.
\textbf{Image-only}: the model receives \emph{only} an image with no
separate text input.
\textbf{Orig.\ script}: questions preserved in their source language
without translation.
\textbf{N$\times$D}: benchmark structure supports factorial analysis
of nation/region and domain effects.
$\dagger$\,subset of items are multimodal.
}
\label{tab:benchmark-comparison}
\small
\resizebox{\textwidth}{!}{%
\begin{tabular}{@{}lrcccccccc@{}}
\toprule
\textbf{Benchmark}
  & \textbf{\#Q}
  & \textbf{\#Lang.}
  & \textbf{\#Scripts}
  & \textbf{\#Domains}
  & \textbf{Multi-modal}
  & \textbf{Image-only}
  & \textbf{Orig.\ script}
  & \textbf{Real exam}
  & \textbf{N$\times$D} \\
\midrule

MMLU~\cite{hendrycks2021measuring}
  & 15,908 & 1 & 1 & 57
  & \xmark & \xmark & --     & \cmark & \xmark \\

MMMU~\cite{yue2023mmmu}
  & 11,550 & 1 & 1 & 30
  & \cmark & \xmark & --     & \cmark & \xmark \\

MATHVISTA~\cite{lu2023mathvista}
  & 6,141  & 1 & 1 & 5
  & \cmark & \xmark & --     & Partial & \xmark \\

AGIEval~\cite{zhong2023agieval}
  & 8,062  & 2 & 2 & 20
  & \xmark & \xmark & \cmark & \cmark & \xmark \\

EXAMS~\cite{hardalov2020exams}
  & 24,000 & 16 & 2 & 24
  & \xmark & \xmark & \cmark & \cmark & \xmark \\

M3Exam~\cite{zhang2023m3exam}
  & 12,317 & 9 & 3 & 3
  & $\dagger$ & \xmark & \cmark & \cmark & \xmark \\

EXAMS-V~\cite{das2024examsv}
  & 20,932 & 11 & 4 & 20
  & \cmark & \cmark & \cmark & \cmark & \xmark \\
\midrule
\textbf{EuraGovExam (Ours)}
  & \textbf{8,000} & \textbf{5+} & \textbf{5} & \textbf{17}
  & \cmark & \cmark & \cmark & \cmark & \cmark \\
\bottomrule
\end{tabular}%
}
\end{table*}

\section{Introduction}

Government examination documents are among the most visually demanding texts that people routinely encounter: a single page may combine dense tables, mathematical expressions, multilingual annotations, and region-specific layout conventions such as vertical Japanese typesetting or mixed-script Devanagari passages.
Millions of civil servants worldwide are assessed through such documents every year, yet the vision-language models (VLMs) that increasingly assist in document analysis have never been systematically evaluated on them.
Existing multimodal benchmarks either supply questions as clean digitized text~\cite{hendrycks2021measuring,zhong2023agieval}, separate images from their textual context~\cite{zhang2023m3exam}, or rely on explicit task-specific prompts that bypass the visual complexity of authentic documents~\cite{lu2023mathvista,yue2023mmmu}.
As a result, it remains unclear how well current VLMs handle the full pipeline of real document understanding---from perceiving complex layouts and scripts to reasoning about the content within them.
This gap is compounded by a pronounced bias toward English and Latin-script content~\cite{hendrycks2021measuring,das2024examsv}, leaving open the question of whether script diversity poses a fundamentally different challenge from domain difficulty.

We introduce \textbf{EuraGovExam}, a multilingual multimodal benchmark sourced from real civil-service examinations across five Eurasian regions: South Korea, Japan, Taiwan, India, and the European Union.
The dataset comprises over 8,000 high-resolution scanned multiple-choice questions spanning 17 academic and administrative domains.
Two design principles distinguish EuraGovExam from prior work.
\textbf{First}, all question content---problem statements, answer choices, tables, figures, and instructions---is embedded within a single image; models receive only a minimal, content-free formatting instruction as text input (\S\ref{sec:protocol}).
\textbf{Second}, questions are preserved in their original language, script, typography, and layout without any translation, OCR extraction, or reformatting, ensuring that the benchmark faithfully reflects the perceptual and linguistic demands of authentic documents.

Our evaluation of 28 VLMs reveals findings that challenge prevailing assumptions about multimodal evaluation.
(i)~\textbf{Region dominates domain}: nation/script factors induce $2.52\times$ greater performance variance than domain factors, with within-model cross-regional gaps reaching 67.9 percentage points---even between regions sharing the same character family (e.g., Japan vs.\ Taiwan).
(ii)~\textbf{Systematic geographic failure patterns}: models achieving over 90\% accuracy on Taiwanese and EU questions can simultaneously fall below 30\% on Japanese questions of comparable format, suggesting that script-specific visual processing constitutes a critical bottleneck.
(iii)~\textbf{Universal failure concentration}: 50 questions (0.6\%) defeat all 28 models, disproportionately concentrated in regions with complex non-Latin scripts and dense visual layouts.

In summary, this paper makes the following contributions:
\begin{itemize}
    \item We present \textbf{EuraGovExam}, a high-fidelity benchmark of 8,000+ image-only exam questions from five Eurasian regions and 17 domains, with a leakage-aware, fully reproducible evaluation protocol.
    \item We demonstrate that \textbf{nation/script is the dominant source of VLM performance variance}, inducing $2.52\times$ greater variance than domain factors---challenging the prevailing domain-centric view of multimodal evaluation.
    \item We provide \textbf{systematic cross-regional and universal failure analysis}, revealing that geographic performance gaps persist across model families and scales, and characterizing the shared visual and linguistic properties of questions that defeat all evaluated models.
\end{itemize}

\section{Related Work}

We position \textbf{EuraGovExam} at the intersection of multimodal question answering, exam-based benchmarks, and multilingual, domain-diverse evaluation datasets. This section reviews prior work across these axes.

\subsection{Multimodal Question Answering}

Multimodal QA has received growing attention with the development of vision-language models. Early efforts such as \textsc{VQA}~\cite{antol2015vqa} and \textsc{GQA}~\cite{hudson2019gqa} focused on everyday scenes, testing spatial and commonsense reasoning. More academic benchmarks like \textsc{TQA}~\cite{kembhavi2017tqa}, \textsc{ScienceQA}~\cite{lu2022learn}, and \textsc{MMMU}~\cite{yue2023mmmu} extend this paradigm to educational domains, incorporating textbook diagrams and subject-specific visuals. \textsc{MathVista}~\cite{lu2023mathvista} further challenges models with mathematical reasoning grounded in visual contexts.

While these benchmarks cover structured content, they are largely English-centric and limited in subject diversity. In contrast, \textsc{EuraGovExam} introduces 17 domains—including law, administration, and philosophy—with fully image-based inputs drawn from real civil service exams across multilingual and regional sources.

\subsection{Exam-Based and Domain-Specific Benchmarks}

Examination-based QA has emerged as a strong proxy for evaluating broad knowledge and reasoning. \textsc{MMLU}~\cite{hendrycks2021measuring} set a standard with 57 subjects across disciplines, but remains text-only and English-based. Domain-specific datasets such as \textsc{HEAD-QA}~\cite{vilares2019headqa} (Spanish medical exams) and \textsc{MedMCQA}~\cite{pal2022medmcqa} (Indian medical entrance tests) highlight the utility of expert-level assessments.

\textsc{AGIEval}~\cite{zhong2023agieval} includes diverse professional exams (e.g., LSAT, SAT, public service) to probe LLM capabilities, yet most remain unimodal. \textsc{EuraGovExam} builds on these works by combining the rigor of real-world exam questions with the complexity of image-based inputs, reflecting authentic visual and linguistic features.

\begin{figure*}
    \centering
    \includegraphics[width=0.9\linewidth]{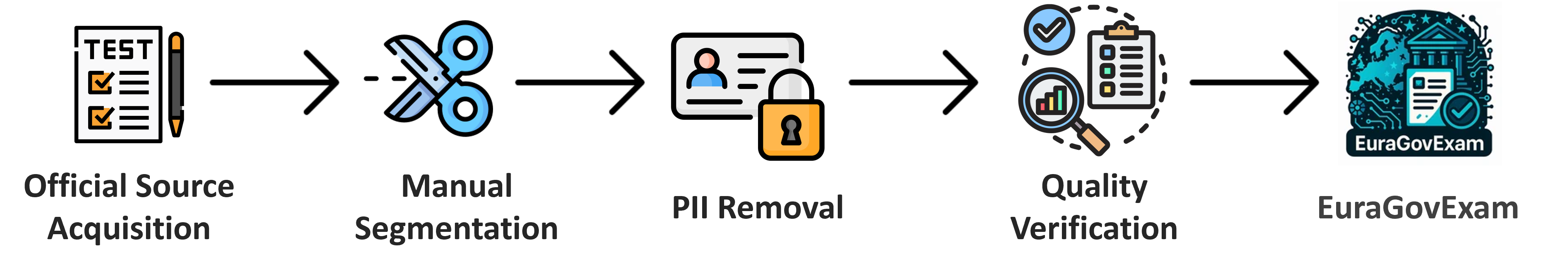}
    \caption{EuraGovExam Dataset Construction Pipeline}
    \label{fig:pipeline}
\end{figure*}

\subsection{Multilingual and Multimodal Exam QA}

Multilingual benchmarks such as \textsc{EXAMS}~\cite{hardalov2020exams}, \textsc{IndoMMLU}~\cite{koto2023indommlu}, and \textsc{CMMLU}~\cite{zeng2023cmmlu} reveal performance gaps in cross-lingual generalization. \textsc{M3Exam}~\cite{zhang2023m3exam} combines multiple languages and modalities, but often presents images and text separately.

\textsc{EXAMS-V}~\cite{das2024examsv} advances this by embedding multimodal questions within unified image inputs, simulating real-world formats. \textsc{EuraGovExam} complements these efforts by covering Eurasian civil exams (e.g., Japan, Korea, Taiwan, India, EU) in a multimodal and multilingual setup, offering a high-fidelity evaluation resource for next-generation vision-language models.

\paragraph{Systematic comparison.}
Table~\ref{tab:benchmark-comparison} positions \textsc{EuraGovExam} against
representative multimodal and exam-based benchmarks along eight axes.
Three properties jointly distinguish our benchmark.
\emph{First}, all question content is embedded within a single image---no
separate text input is provided to the model, unlike benchmarks that supply
OCR-extracted text or image--text pairs.
\emph{Second}, questions are preserved in their original script and layout
without translation, covering five distinct script families (Hangul, Kanji/Kana,
Traditional Chinese, Devanagari, and Latin) within a single evaluation suite.
\emph{Third}, the combination of 17 professional domains with five regional
sources enables factorial analysis of domain versus nation/script effects on
VLM performance---an analysis axis unavailable in any prior benchmark.


\begin{figure*}[t]
    \centering
    \begin{subfigure}[t]{0.48\textwidth}
        \centering
        \includegraphics[width=\linewidth]{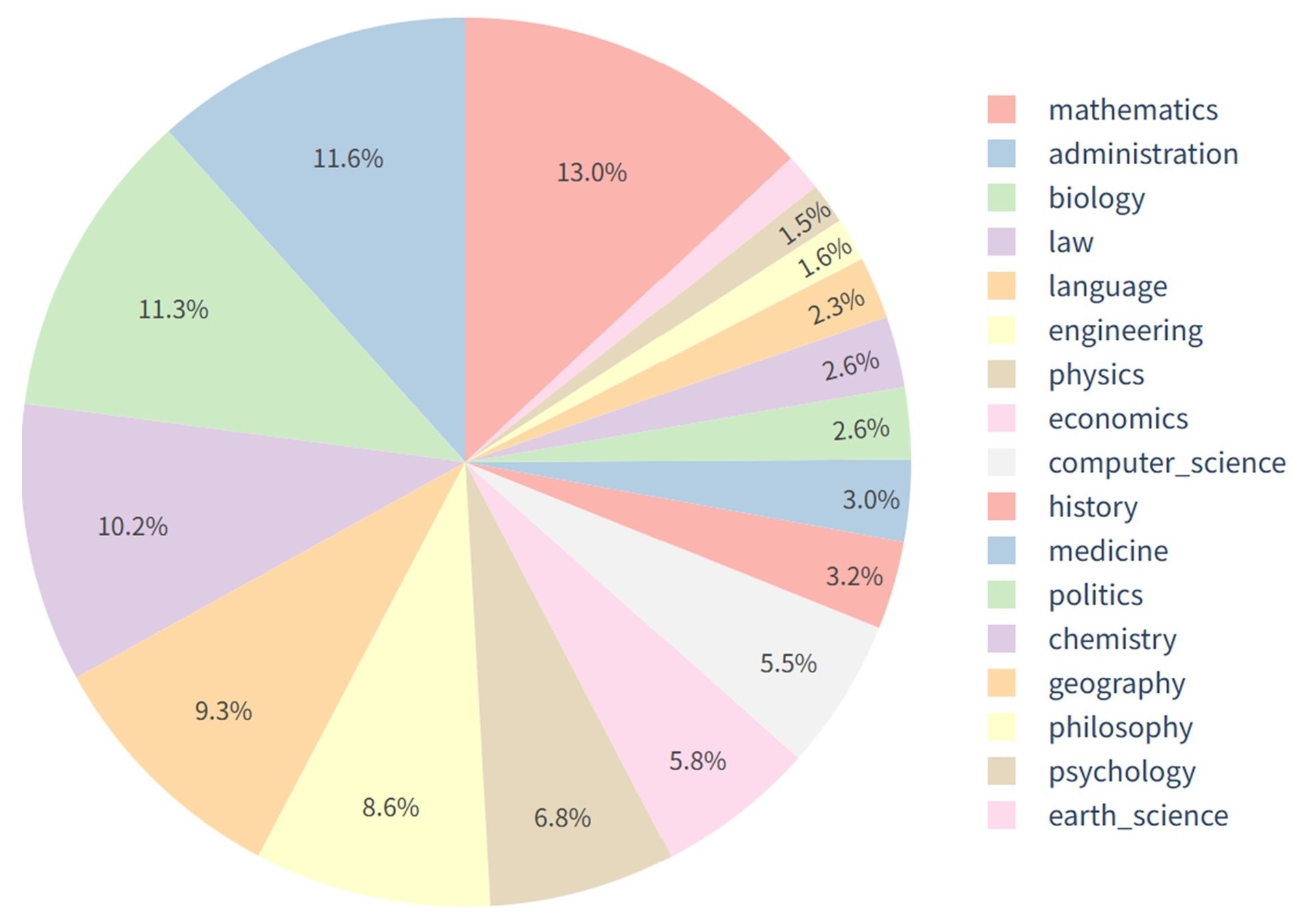}
        \caption{Distribution of tasks in the EuraGovExam dataset across different reasoning types.}
        \label{fig:dist_task}
    \end{subfigure}
    \hfill
    \begin{subfigure}[t]{0.43\textwidth}
        \centering
        \includegraphics[width=\linewidth]{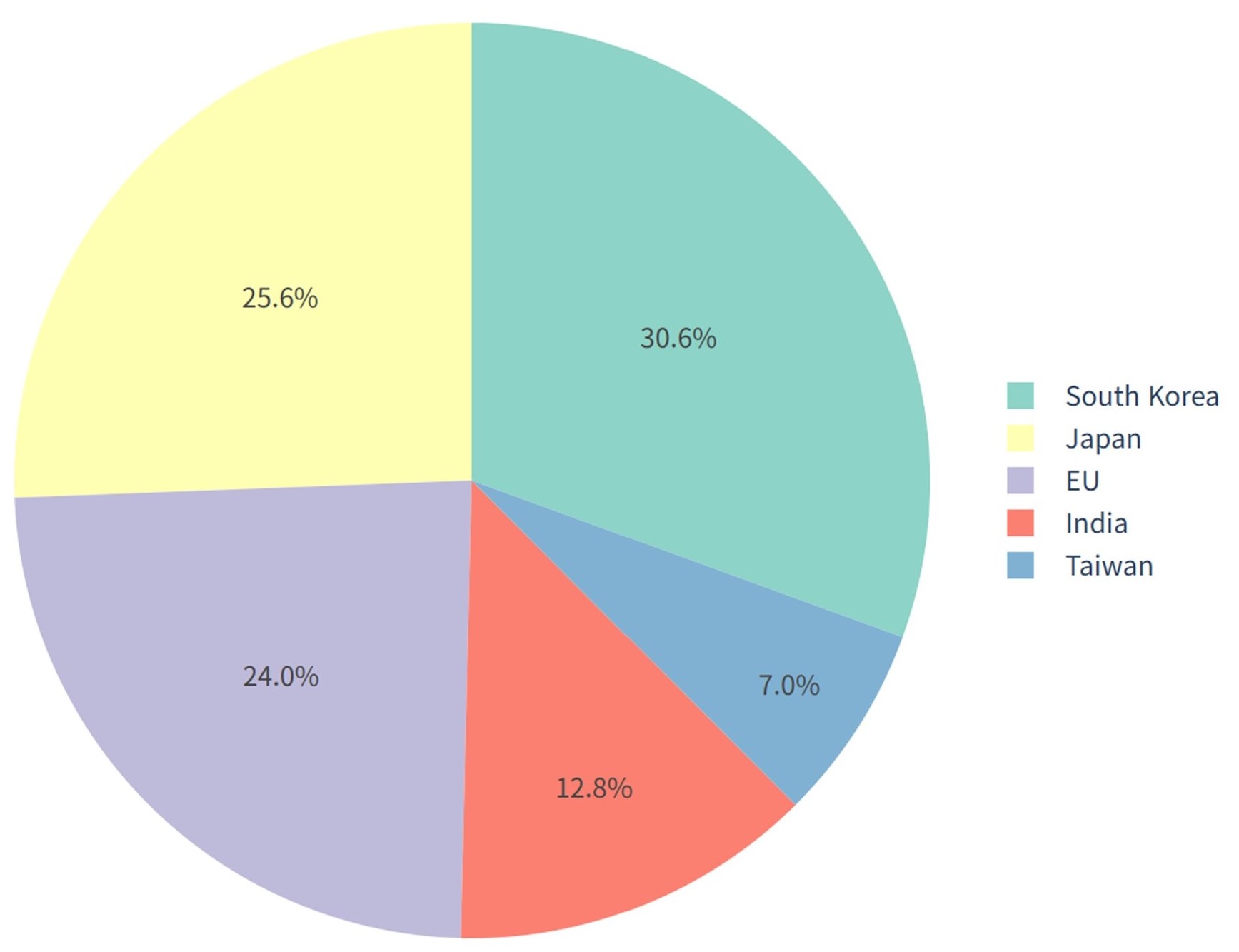}
        \caption{Distribution of the EuraGovExam dataset by nation of origin.}
        \label{fig:dist_nation}
    \end{subfigure}
    \caption{Distribution of the EuraGovExam dataset.}
    \label{fig:distribution}
\end{figure*}

\section{Dataset Construction}
\label{sec:dataset_construction}

EuraGovExam comprises \textbf{real multiple-choice questions} collected from public-sector examinations (national and regional civil service recruitment tests) across \textbf{five Eurasian regions}.
Unlike conventional multimodal benchmarks that provide ``image + separate text question'' pairs,
our dataset is designed such that \textbf{all question content---passages, answer choices, tables, figures, mathematical expressions, and multilingual annotations---exists solely within a single image}.
This design enables EuraGovExam to evaluate VLMs' \emph{document layout understanding} and \emph{vision-language reasoning} capabilities under conditions that closely approximate real examination settings.

\subsection{Data Sources and Provenance}
\label{subsec:data_sources}

EuraGovExam draws from \textbf{officially released materials} across five regions:
\textbf{South Korea}, \textbf{Japan}, \textbf{Taiwan}, \textbf{India}, and the \textbf{European Union (EU)}.
We adhere to \textit{institutional legitimacy} and \textit{legal compliance} as primary principles,
using \textbf{only materials provided by government portals or official agencies}.

\paragraph{Source provenance.}
\begin{itemize}
    \item \textbf{South Korea}: Publicly released past examination questions from national civil service tests.
    \item \textbf{Japan}: Past examination questions and sample problems released by the \textit{National Personnel Authority}.
    \item \textbf{Taiwan}: Archived public examination items from the \textit{Ministry of Examination}.
    \item \textbf{India}: Previous question papers publicly released by the \textit{Union Public Service Commission} (UPSC).
    \item \textbf{European Union}: Official \textbf{sample and exemplar questions} from \textit{EPSO} (European Personnel Selection Office).
    While actual administered examinations are generally not disclosed, EPSO provides official sample questions with equivalent format and calibrated difficulty levels.
\end{itemize}

\subsection{Construction Pipeline}
\label{subsec:construction_pipeline}

Figure~\ref{fig:pipeline} summarizes the data construction pipeline.
The primary objective is to preserve \textbf{visual and structural fidelity}.
To this end, we eschew OCR-based automatic extraction in favor of \textbf{fully manual segmentation}.

\paragraph{Step 1: Document acquisition.}
We collect original documents from official portals (predominantly PDF scans or digitally distributed versions),
preserving the original resolution and rendering whenever possible.

\paragraph{Step 2: Manual question segmentation.}
Trained annotators perform individual cropping at the question level from source documents.
Questions are segmented without alteration---only separated by item boundaries.
We exclude OCR to prevent (i) semantic corruption due to recognition errors, (ii) loss of tables, figures, mathematical expressions, and multi-column layouts,
and (iii) potential omissions in mixed-language text or special characters.

\paragraph{Step 3: Format preservation.}
All questions are retained in their original form \textbf{without translation, paraphrasing, or rewriting}.
Specifically, we preserve the original language, script, font and typography, layout, and all graphical elements including tables and figures,
thereby reflecting the same input distribution encountered in actual examination settings.

\paragraph{Step 4: Quality verification.}
Each question undergoes a two-stage review process based on the following criteria: (i) answer label consistency, (ii) crop completeness (absence of truncated tables, figures, or answer choices),
and (iii) resolution and legibility (absence of blur or compression artifacts).
Items failing verification are either re-segmented or excluded.

\subsection{Privacy, Ethics, and Licensing}
\label{subsec:privacy_licensing}

\paragraph{Privacy.}
EuraGovExam does not contain individual-level test-taker data (responses or scores).
When residual PII (e.g., candidate numbers or personal identification marks) may be present in question images,
annotators \textbf{manually remove} such information.

\paragraph{Licensing and reuse.}
The dataset is intended primarily for \textbf{non-commercial research purposes}, respecting the disclosure and reuse policies of source institutions.
Japanese materials fall under PDL 1.0 (reuse permitted with attribution)~\cite{japan_pdl10};
Taiwanese materials are governed by OGDL-Taiwan-1.0 (redistribution and modification permitted with attribution and adherence to exception clauses)~\cite{taiwan_ogdl10};
EU Careers/EPSO content is generally subject to CC BY 4.0 (attribution and indication of modifications required), with potential third-party exceptions~\cite{eu_legal_notice};
Indian UPSC materials are provided by segmenting original pages at the question level \textbf{without content modification}~\cite{upsc_website_policy}.
Korean materials follow the policies of their respective administering and publishing agencies,
with KOGL (Korea Open Government License) conditions applied where such attribution is confirmed~\cite{kogl_license}.

As jurisdictional constraints may vary, users intending purposes beyond research (e.g., commercial redistribution) should independently verify applicable regional regulations.

\subsection{Problem Format and Visual Structure}
\label{subsec:problem_format}

Each item is presented as an \textbf{image}, with question content containing diverse document structures:
tables, graphs, multi-column layouts, mixed-language annotations, special symbols, and mathematical expressions.
Answers follow a multiple-choice format (4 or 5 options), requiring models to identify answer choices from the image,
perform the required operations (calculation, comparison, reasoning, or comprehension), and output the correct answer.

Questions frequently demand \textbf{document layout-grounded reasoning}:
(i) referencing specific rows or columns in tables to compare values,
(ii) cross-referencing when passage language differs from table or figure labels,
and (iii) combining mathematical expressions or diagram interpretation with textual instructions.
These characteristics constitute a core difficulty axis distinguishing EuraGovExam from conventional text-centric benchmarks.

\section{Evaluation Kit and Protocol}
\label{sec:protocol}

This section defines the \textbf{standardized evaluation kit} and
\textbf{reproducible evaluation protocol} for EuraGovExam.
We fix (i) input/output formats, (ii) inference rules, and (iii) scoring rules
to ensure that comparisons across models and platforms are conducted under \textbf{identical conditions}.

\subsection{Image-Only Setting}
\label{subsec:image_only}

EuraGovExam adopts an \textbf{image-only} setting. The model input consists solely of a single question image,
with all question text (passages, answer choices, and instructions) existing only within the image.
This setting adheres to the following principles.

\paragraph{Principle 1: No external text input.}
The text provided to the model contains no question content.
Specifically, no separate question text, JSON metadata, or OCR results are supplied as input.

\paragraph{Principle 2: No OCR or external tools.}
During evaluation, the use of OCR engines, external search, calculators, translators, or any other tools is prohibited.
Models must recognize and reason about questions in an \textbf{end-to-end} manner.

\paragraph{Principle 3: Standardized minimal instruction only.}
All models receive only an identical format instruction (containing no question information):
\begin{quote}\small
\texttt{You are solving a multiple-choice exam question shown in the image.
At the very end, provide the final answer in exactly this format:
The answer is X. (e.g., The answer is B.)}
\end{quote}
This design minimizes variance due to prompt engineering and ensures fair comparison.

\subsection{Inference Rules}
\label{subsec:inference_rules}

\paragraph{Input preprocessing.}
Input images are used at their original resolution by default.
When model-specific maximum input size constraints apply, resizing is performed while \textbf{preserving the aspect ratio}.
Resizing rules are specified in the evaluation script and applied consistently for each model under identical conditions.

\paragraph{Generation budget.}
We fix the maximum generation length $L_{\max}$ for all models,
with the answer restricted to a single final line in the format \texttt{The answer is X.}
Both $L_{\max}$ and stop rules are specified in the evaluation script.

\paragraph{Answer extraction.}
All models must provide their answer in the final line using the format \texttt{The answer is X.}
The evaluation script parses $X \in \{A, B, C, D, E\}$ using regular expressions.
Format violations, multiple answers, or missing outputs are scored as incorrect,
with all rules publicly documented to eliminate scoring ambiguity.

\begin{table*}[t]
  \centering

  \caption{%
\textbf{EuraGovExam performance.} %
Icons denote regions %
(\protect\includegraphics[height=1em]{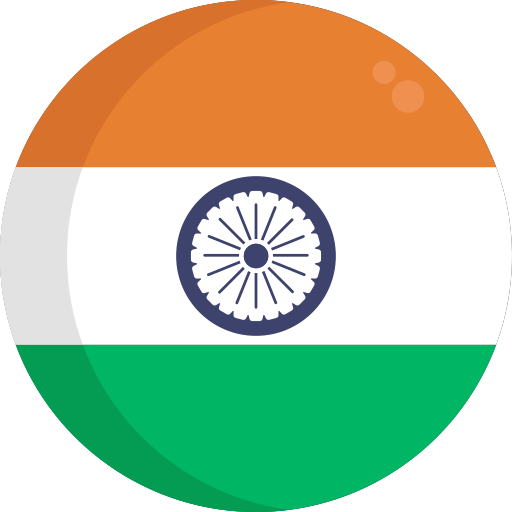}\,India, %
\protect\includegraphics[height=1em]{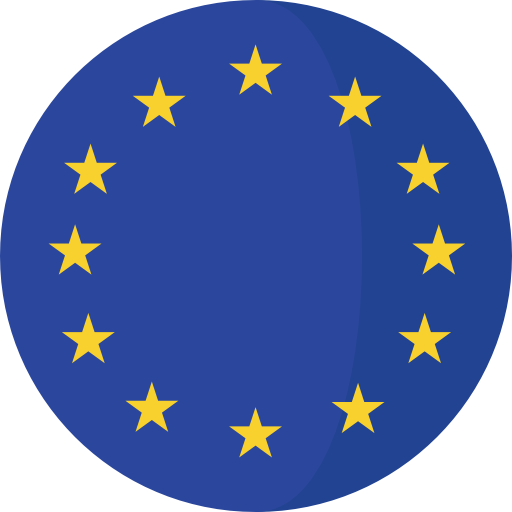}\,EU, %
\protect\includegraphics[height=1em]{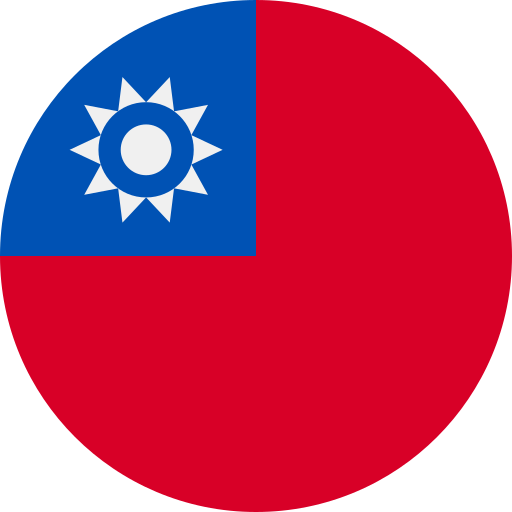}\,Taiwan, %
\protect\includegraphics[height=1em]{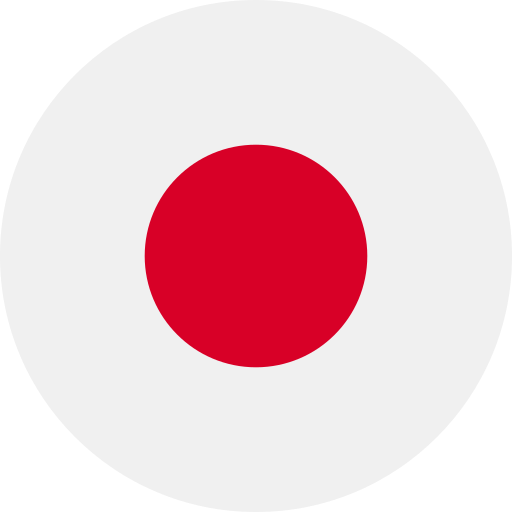}\,Japan, %
\protect\includegraphics[height=1em]{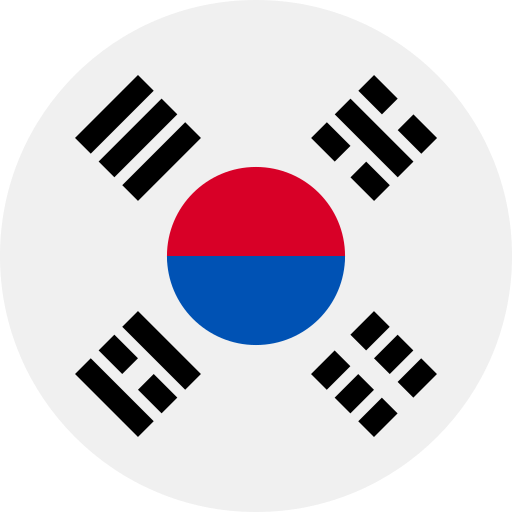}\,S.\,Korea) %
and domains %
(\protect\includegraphics[height=1em]{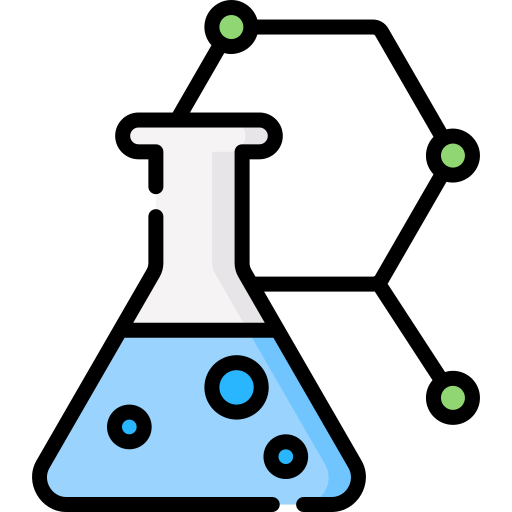}\,Chem, %
\protect\includegraphics[height=1em]{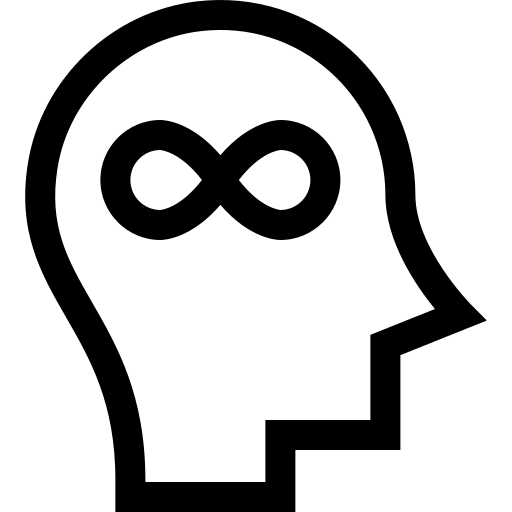}\,Phil, %
\protect\includegraphics[height=1em]{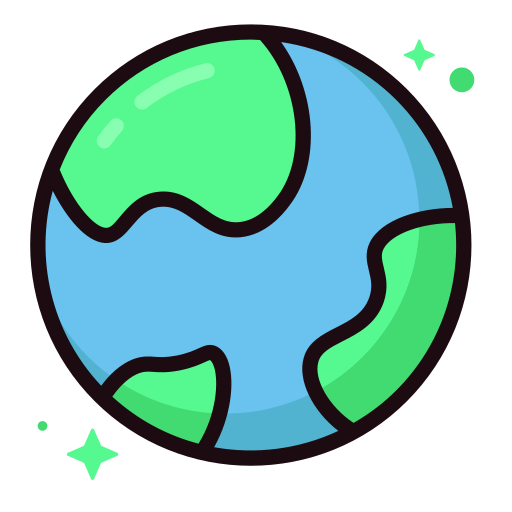}\,Earth, %
\protect\includegraphics[height=1em]{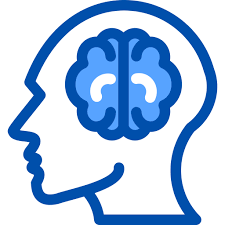}\,Psych, %
\protect\includegraphics[height=1em]{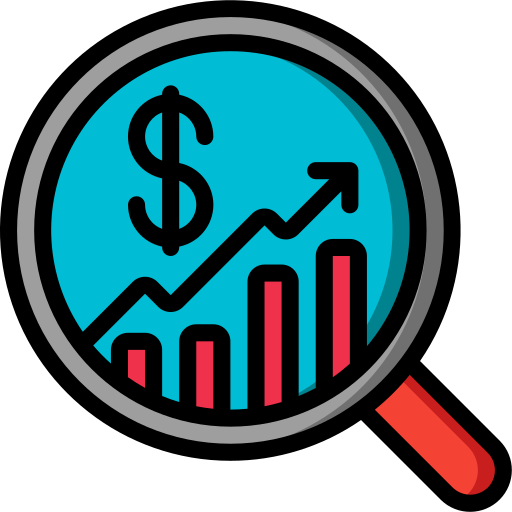}\,Econ, %
\protect\includegraphics[height=1em]{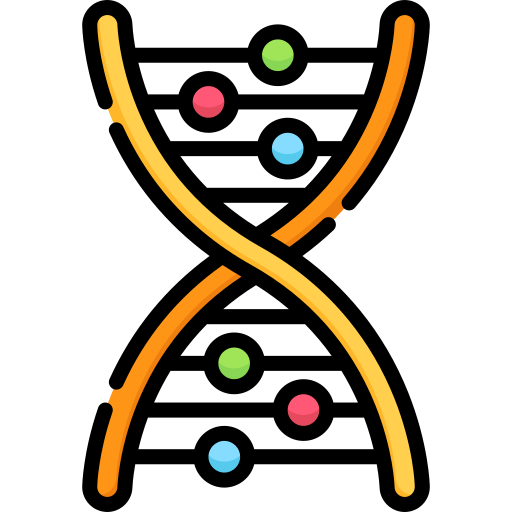}\,Bio, %
\protect\includegraphics[height=1em]{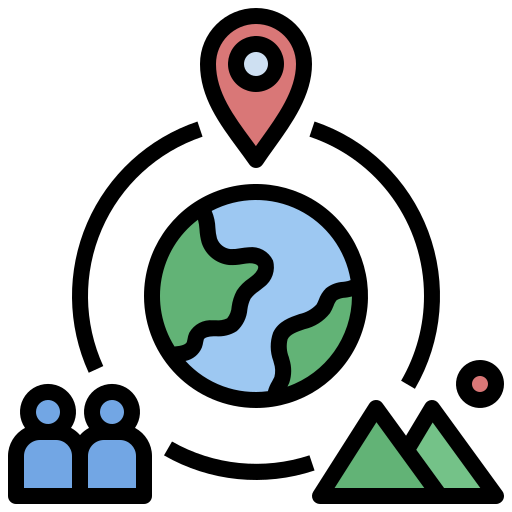}\,Geo, %
\protect\includegraphics[height=1em]{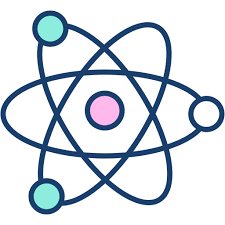}\,Phys, %
\protect\includegraphics[height=1em]{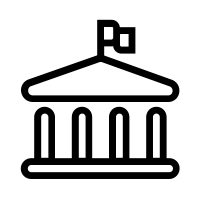}\,Pol, %
\protect\includegraphics[height=1em]{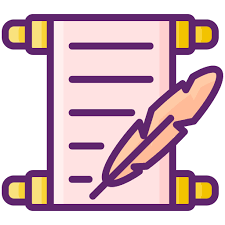}\,Hist, %
\protect\includegraphics[height=1em]{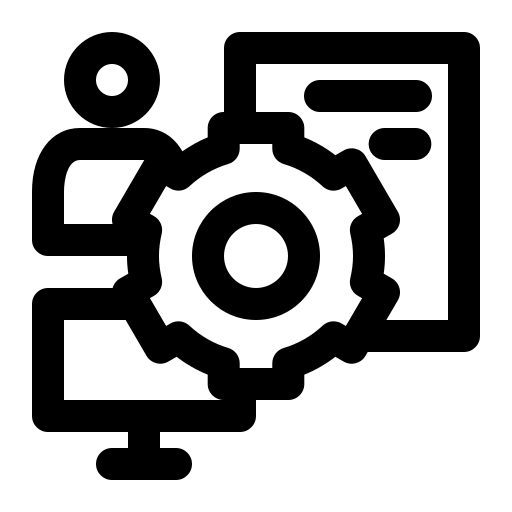}\,Admin, %
\protect\includegraphics[height=1em]{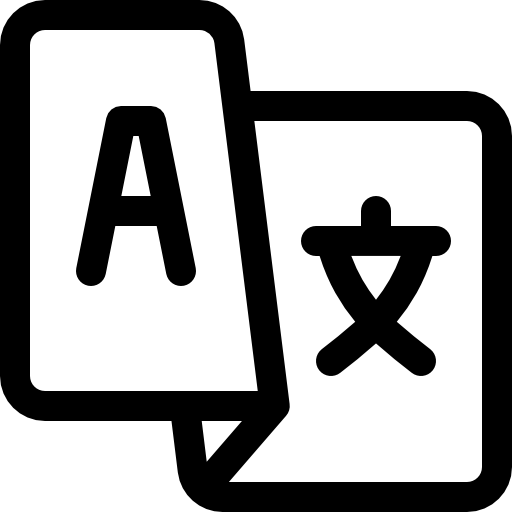}\,Lang, %
\protect\includegraphics[height=1em]{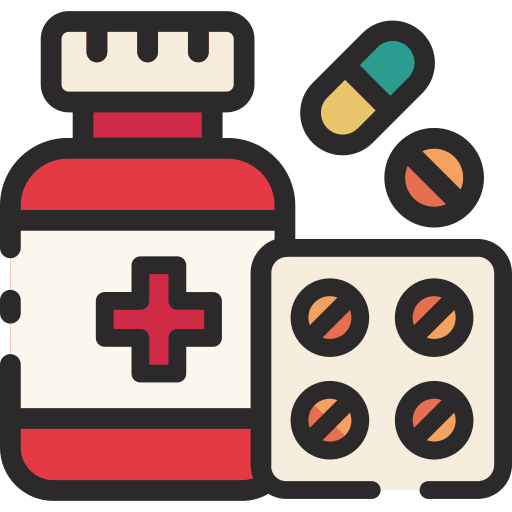}\,Med, %
\protect\includegraphics[height=1em]{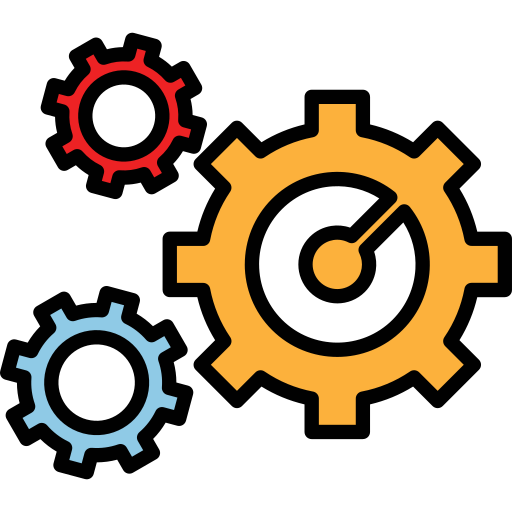}\,Eng, %
\protect\includegraphics[height=1em]{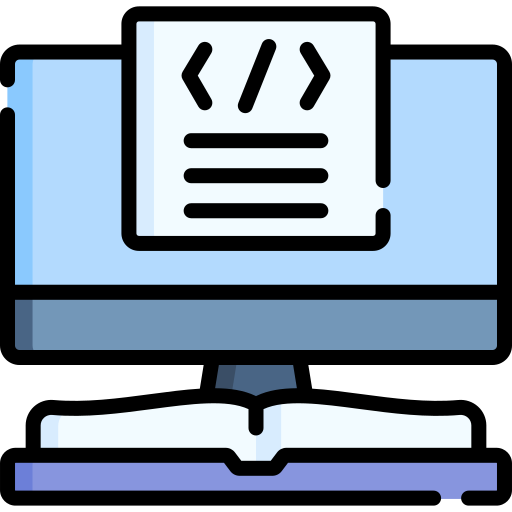}\,CS, %
\protect\includegraphics[height=1em]{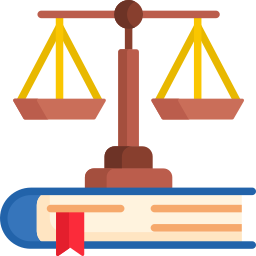}\,Law, %
\protect\includegraphics[height=1em]{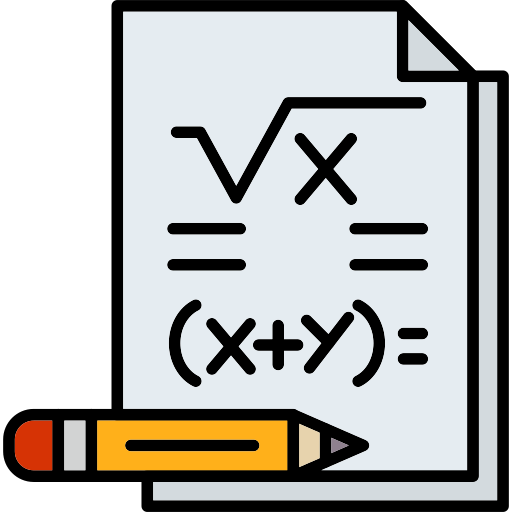}\,Math). %
All models evaluated under the image-only protocol (\S\ref{sec:protocol})
with greedy decoding and single inference.
}

\label{tab:main_results}
\resizebox{\textwidth}{!}{%
  \begin{tabular}{l*{28}{c}}
    \toprule
    \multirow{2}{*}{\textbf{Models}}
      & \multirow{2}{*}{\textbf{Overall}}
      & \multicolumn{5}{c}{\textbf{Nation}}
      & \multicolumn{17}{c}{\textbf{Task}} \\
    \cmidrule(lr){3-7} \cmidrule(lr){8-24}
      &
      & {\includegraphics[height=1.5em]{figure/icon/india.png}}
      & {\includegraphics[height=1.5em]{figure/icon/EU.png}}
      & {\includegraphics[height=1.5em]{figure/icon/Taiwan.png}}
      & {\includegraphics[height=1.5em]{figure/icon/Japan.png}}
      & {\includegraphics[height=1.5em]{figure/icon/South_Korea.png}}
      & {\includegraphics[height=1.5em]{figure/icon/chemistry.png}}
      & {\includegraphics[height=1.5em]{figure/icon/philosophy.png}}
      & {\includegraphics[height=1.5em]{figure/icon/earth.png}}
      & {\includegraphics[height=1.5em]{figure/icon/psychology.png}}
      & {\includegraphics[height=1.5em]{figure/icon/economics.png}}
      & {\includegraphics[height=1.5em]{figure/icon/biology.png}}
      & {\includegraphics[height=1.5em]{figure/icon/geography.png}}
      & {\includegraphics[height=1.5em]{figure/icon/physics.png}}
      & {\includegraphics[height=1.5em]{figure/icon/politics.png}}
      & {\includegraphics[height=1.5em]{figure/icon/history.png}}
      & {\includegraphics[height=1.5em]{figure/icon/administration.png}}
      & {\includegraphics[height=1.5em]{figure/icon/language.png}}
      & {\includegraphics[height=1.5em]{figure/icon/medicine.png}}
      & {\includegraphics[height=1.5em]{figure/icon/engineering.png}}
      & {\includegraphics[height=1.5em]{figure/icon/computer_science.png}}
      & {\includegraphics[height=1.5em]{figure/icon/law.png}}
      & {\includegraphics[height=1.5em]{figure/icon/mathematics.png}} \\[-0.2em]

    \cmidrule(lr){1-24}
    \multicolumn{24}{c}{{Closed-source Models}} \\
    \cmidrule(lr){1-24}
    o3~\cite{openai2025o4mini}
      & 84.26 & 68.64 & 84.49 & 93.72 & 82.37 & 90.06
      & 85.51 & 90.55 & 73.83 & 92.44 & 87.20 & 89.23 & 74.03 & 81.85 & 79.90 & 86.33 & 81.53 & 91.49 & 91.18 & 84.16 & 90.93 & 78.08 & 78.98 \\
    o4-mini~\cite{openai2025o4mini}
      & 79.40 & 63.38 & 76.95 & 92.29 & 82.52 & 82.49
      & 85.02 & 85.83 & 75.70 & 85.71 & 82.86 & 83.02 & 64.09 & 89.63 & 73.21 & 67.97 & 71.21 & 84.19 & 88.24 & 81.98 & 92.52 & 63.92 & 80.61 \\
    GPT-4o~\cite{hurst2024gpt}
      & 42.04 & 40.99 & 63.73 & 66.66 & 25.97 & 33.25
      & 50.72 & 36.22 & 17.76 & 29.41 & 31.89 & 42.73 & 37.02 & 33.52 & 34.93 & 35.55 & 47.48 & 49.73 & 50.00 & 33.72 & 58.96 & 42.00 & 43.38 \\
    GPT-4.1~\cite{openai2025gpt41}
      & 54.73 & 48.10 & 66.44 & 72.58 & 48.10 & 54.23
      & 68.12 & 55.12 & 33.64 & 46.22 & 46.42 & 53.50 & 48.07 & 54.63 & 45.93 & 48.83 & 54.57 & 60.14 & 69.75 & 50.00 & 73.47 & 47.54 & 58.06 \\
    GPT-4.1-mini~\cite{openai2025gpt41}
      & 56.27 & 46.34 & 63.58 & 79.03 & 43.84 & 59.92
      & 65.22 & 72.44 & 34.58 & 52.10 & 51.84 & 58.93 & 37.02 & 58.70 & 47.37 & 39.84 & 54.14 & 61.62 & 72.69 & 51.16 & 75.74 & 51.23 & 56.24 \\
    GPT-5-nano~\cite{openai2025gpt5nano}
      & 67.58 & 52.19 & 63.68 & 83.87 & 68.07 & 72.97
      & 73.91 & 74.80 & 50.47 & 73.95 & 69.20 & 73.70 & 51.38 & 73.52 & 57.42 & 50.00 & 63.59 & 61.08 & 81.51 & 68.17 & 83.67 & 57.14 & 72.46 \\
    GPT-5~\cite{openai2025gpt5}
      & 85.80 & 68.35 & 83.87 & 94.80 & 87.50 & 91.17
      & 86.96 & 92.91 & 74.77 & 94.96 & 89.37 & 90.01 & 70.72 & 87.96 & 79.90 & 86.33 & 83.67 & 90.27 & 93.28 & 88.81 & 92.06 & 81.03 & 78.22 \\
    GPT-5.2~\cite{openai2025gpt52}
      & 69.94 & 53.46 & 68.05 & 78.67 & 73.83 & 73.09
      & 78.26 & 79.53 & 61.68 & 70.59 & 71.80 & 67.48 & 59.12 & 78.52 & 67.94 & 61.33 & 68.53 & 61.35 & 81.09 & 75.00 & 85.94 & 63.92 & 68.52 \\
    Gemini-3-pro-preview~\cite{google2025gemini3propreview}
      & 68.53 & 68.94 & 92.82 & 97.49 & 29.59 & 75.26
      & 75.36 & 78.74 & 35.51 & 66.39 & 62.04 & 77.03 & 62.98 & 28.89 & 64.11 & 74.22 & 77.23 & 86.89 & 88.66 & 55.23 & 73.47 & 71.67 & 64.88 \\
    Gemini-3-flash-preview~\cite{google2025gemini3flashpreview}
      & 75.28 & 51.70 & 89.54 & 97.31 & 56.69 & 84.50
      & 77.78 & 85.04 & 44.86 & 71.43 & 65.08 & 78.36 & 57.46 & 63.52 & 62.68 & 71.09 & 79.16 & 85.81 & 92.86 & 72.38 & 85.94 & 76.48 & 73.22 \\
    Gemini-2.5-pro~\cite{comanici2025gemini}
      & 86.99 & 69.23 & 88.08 & 95.51 & 87.59 & 91.12
      & 87.92 & 93.70 & 85.98 & 95.80 & 88.29 & 88.90 & 74.03 & 84.26 & 79.43 & 87.50 & 83.57 & 91.49 & 95.80 & 86.19 & 92.06 & 84.24 & 86.28 \\
    Gemini-2.5-flash~\cite{comanici2025gemini}
      & 68.33 & 62.32 & 83.30 & 92.65 & 51.46 & 67.65
      & 72.95 & 77.17 & 48.60 & 66.39 & 55.97 & 73.36 & 54.70 & 59.26 & 62.20 & 66.02 & 68.53 & 75.68 & 83.19 & 64.83 & 75.28 & 63.18 & 73.13 \\
    Gemini-2.5-flash-lite~\cite{comanici2025gemini}
      & 25.94 & 12.95 & 54.37 & 73.29 & 07.13 & 13.99
      & 22.71 & 18.11 & 7.48 & 17.65 & 16.49 & 40.40 & 23.20 & 5.74 & 23.45 & 16.80 & 37.27 & 37.03 & 42.02 & 13.23 & 33.33 & 34.36 & 12.76 \\
    Claude-Sonnet-4~\cite{anthropic2025claude4}
      & 63.29 & 62.51 & 76.43 & 87.28 & 45.85 & 62.41
      & 67.63 & 70.87 & 40.19 & 64.71 & 56.83 & 68.81 & 55.80 & 51.67 & 63.16 & 66.02 & 64.12 & 72.30 & 78.99 & 56.40 & 69.39 & 60.84 & 61.61 \\

    \cmidrule(lr){1-24}
    \multicolumn{24}{c}{{Open-source Models}} \\
    \cmidrule(lr){1-24}
    Llama-3.2-11B-Vision~\cite{meta2024llama3}
      & 12.75 & 13.82 & 20.08 & 23.65 & 10.06 & 6.29
      & 10.63 & 7.87 & 15.89 & 7.56 & 10.63 & 16.09 & 23.20 & 11.11 & 13.88 & 11.72 & 15.79 & 13.11 & 13.87 & 8.87 & 15.42 & 16.50 & 6.44 \\
    Qwen2-VL-2B-Instruct~\cite{wang2024qwen2}
      & 25.54 & 15.09 & 35.90 & 33.33 & 18.21 & 26.13
      & 24.64 & 19.68 & 15.89 & 27.73 & 22.99 & 27.08 & 17.68 & 19.63 & 22.97 & 18.36 & 33.40 & 34.32 & 25.63 & 22.97 & 26.30 & 28.82 & 19.19 \\
    Qwen2-VL-7B-Instruct~\cite{wang2024qwen2}
      & 31.38 & 27.36 & 47.19 & 52.51 & 15.57 & 29.08
      & 28.50 & 33.86 & 12.15 & 26.89 & 26.03 & 37.51 & 30.39 & 16.11 & 29.67 & 25.39 & 41.78 & 41.89 & 37.82 & 20.93 & 41.04 & 37.81 & 20.73 \\
    Qwen2.5-VL-7B-Instruct~\cite{bai2025qwen2}
      & 32.30 & 26.29 & 45.89 & 46.95 & 21.88 & 29.53
      & 33.33 & 30.71 & 18.69 & 25.21 & 25.16 & 35.18 & 30.39 & 25.00 & 24.40 & 26.95 & 36.52 & 43.51 & 36.13 & 26.02 & 41.72 & 34.48 & 28.02 \\
    Phi-3.5-vision-instruct~\cite{abdin2024phi3}
      & 15.67 & 15.48 & 19.56 & 15.77 & 14.64 & 13.53
      & 18.36 & 11.02 & 10.28 & 16.81 & 14.32 & 14.21 & 19.34 & 13.52 & 17.70 & 16.80 & 19.01 & 15.54 & 13.45 & 15.41 & 16.55 & 18.60 & 12.96 \\
    Qwen2-VL-72B-Instruct~\cite{wang2024qwen2}
      & 44.65 & 35.93 & 62.07 & 74.73 & 30.37 & 39.71
      & 52.66 & 41.73 & 26.17 & 38.66 & 34.27 & 48.72 & 37.57 & 37.41 & 35.41 & 31.25 & 51.99 & 53.38 & 50.84 & 35.61 & 58.28 & 48.03 & 40.60 \\
    InternVL2.5-38B-MPO~\cite{wang2024enhancing}
      & 39.34 & 19.38 & 52.91 & 56.81 & 31.49 & 39.63
      & 38.16 & 30.71 & 14.02 & 34.45 & 27.12 & 41.18 & 24.86 & 36.48 & 35.89 & 26.56 & 44.04 & 49.73 & 39.50 & 34.74 & 56.24 & 40.02 & 39.16 \\
    Ovis2-8B~\cite{lu2024ovis}
      & 28.31 & 25.02 & 40.63 & 34.77 & 17.58 & 27.53
      & 30.43 & 31.50 & 17.76 & 21.85 & 23.64 & 29.97 & 29.83 & 17.41 & 25.84 & 24.22 & 36.09 & 37.30 & 26.05 & 21.51 & 34.92 & 31.16 & 23.51 \\
    Ovis2-16B~\cite{lu2024ovis}
      & 32.73 & 20.06 & 45.32 & 51.61 & 26.71 & 28.88
      & 33.82 & 25.98 & 19.63 & 21.85 & 22.99 & 35.29 & 25.41 & 32.59 & 29.67 & 24.22 & 39.21 & 42.97 & 35.29 & 26.16 & 38.78 & 30.05 & 32.25 \\
    Ovis2-32B~\cite{lu2024ovis}
      & 35.50 & 22.49 & 51.61 & 54.48 & 28.42 & 29.90
      & 32.37 & 30.71 & 15.89 & 23.53 & 25.60 & 37.62 & 27.62 & 36.11 & 30.14 & 18.75 & 41.25 & 45.95 & 40.34 & 24.71 & 43.99 & 38.05 & 36.76 \\
    llama3-llava-next-8b~\cite{li2024llavanext}
      & 14.28 & 7.89 & 16.65 & 17.92 & 12.30 & 15.91
      & 20.77 & 19.69 & 7.48 & 15.97 & 12.80 & 14.21 & 12.71 & 14.07 & 9.09 & 15.63 & 17.29 & 14.86 & 16.81 & 11.77 & 12.93 & 16.01 & 11.80 \\
    llava-1.5-13b~\cite{liu2024improved}
      & 18.99 & 16.65 & 21.59 & 17.92 & 19.48 & 17.75
      & 15.94 & 16.54 & 18.69 & 19.33 & 19.52 & 17.31 & 23.76 & 19.44 & 19.62 & 16.41 & 21.80 & 18.51 & 21.43 & 18.31 & 17.23 & 17.00 & 20.54 \\
    llava-1.5-7b~\cite{liu2024improved}
      & 12.44 & 12.76 & 13.22 & 12.37 & 11.96 & 12.11
      & 11.59 & 14.17 & 11.21 & 6.72 & 13.02 & 11.88 & 10.50 & 12.96 & 10.53 & 12.11 & 13.75 & 13.11 & 13.03 & 12.21 & 14.74 & 12.19 & 11.52 \\
    LLaVA-NeXT-Video-7B-DPO-hf~\cite{zhang2024video}
      & 7.21 & 10.22 & 6.56 & 9.68 & 7.18 & 5.93
      & 9.66 & 2.36 & 5.61 & 5.88 & 7.81 & 6.99 & 8.84 & 7.22 & 7.66 & 9.38 & 5.26 & 5.95 & 6.72 & 7.56 & 5.44 & 6.90 & 10.17 \\

    \bottomrule
  \end{tabular}

}

\end{table*}

\subsection{Metrics and Breakdown}
\label{subsec:metrics}

\paragraph{Primary metric.}
The primary evaluation metric is accuracy:
\begin{equation}
\mathrm{Accuracy} = \frac{1}{N}\sum_{i=1}^{N}\mathbb{1}[\hat{y}_i = y_i],
\label{eq:acc}
\end{equation}
where $\hat{y}_i$ denotes the predicted answer, $y_i$ the ground-truth answer, and $N$ the number of questions.

\paragraph{Random baseline.}
The random baseline for 4-choice questions is 25.0\% and for 5-choice questions is 20.0\%.
Based on the dataset composition, the weighted average random baseline is \textbf{23.7\%}.
Models performing below this threshold are interpreted as exhibiting systematic parsing errors or fundamental failures in question comprehension.

\paragraph{Breakdown axes.}
In addition to overall accuracy, we report results along three breakdown axes:
\begin{itemize}
    \item \textbf{By nation}: Accuracy across five regions (South Korea, Japan, Taiwan, India, EU).
    \item \textbf{By domain}: Accuracy across 17 domains (e.g., mathematics, administration, biology).
    \item \textbf{Failure overlap}: Identification of question sets where multiple models commonly fail, enabling diagnosis of \emph{systematic bottlenecks}.
\end{itemize}

\subsection{Reproducibility Statement}
\label{subsec:reproducibility}

\begin{tcolorbox}[
    enhanced,
    breakable,
    colback=black!90,
    colframe=black,
    coltitle=white,
    fonttitle=\bfseries\ttfamily\small,
    fontupper=\ttfamily\small\color{green!70!white},
    title={\textcolor{red!70}{$\bullet$} \textcolor{yellow!70}{$\bullet$} \textcolor{green!70}{$\bullet$}},
    boxrule=0pt,
    arc=4pt,
    left=6pt, right=6pt, top=4pt, bottom=4pt,
    shadow={2pt}{-2pt}{0pt}{black!50}
]
\textcolor{gray!60}{\# Full benchmark evaluation}\\
\textcolor{cyan!80}{\$} python evaluate.py \textbackslash\\
\hspace{2em}--model \textcolor{yellow!80}{\{MODEL\_NAME\}} \textbackslash\\
\hspace{2em}--split \textcolor{yellow!80}{test} \textbackslash\\
\hspace{2em}--setting \textcolor{yellow!80}{image-only}\\[6pt]
\textcolor{gray!60}{\# Filter by nation (e.g., Japan only)}\\
\textcolor{cyan!80}{\$} python evaluate.py --model \textcolor{yellow!80}{gpt-4o} --nation \textcolor{yellow!80}{japan}\\[6pt]
\textcolor{gray!60}{\# Filter by domain (e.g., mathematics only)}\\
\textcolor{cyan!80}{\$} python evaluate.py --model \textcolor{yellow!80}{gpt-4o} --domain \textcolor{yellow!80}{mathematics}\\[6pt]
\textcolor{gray!60}{\# Combine filters (nation + domain)}\\
\textcolor{cyan!80}{\$} python evaluate.py --model \textcolor{yellow!80}{gpt-4o} --nation \textcolor{yellow!80}{korea} --domain \textcolor{yellow!80}{law}
\end{tcolorbox}

To ensure reproducibility of EuraGovExam results, we provide a \textbf{standardized evaluation script}, \textbf{fixed data splits},
\textbf{fixed inference rules} (decoding, preprocessing, and parsing), and \textbf{experiment logging templates}.
All main results can be reproduced with a single command.

\noindent The evaluation script supports the following filtering options:
\texttt{--nation} (five regions: \texttt{korea}, \texttt{japan}, \texttt{taiwan}, \texttt{india}, \texttt{eu}) and
\texttt{--domain} (17 subjects).
Multiple filters can be combined for fine-grained analysis, with results automatically saved in JSON format.

\section{Benchmark Results}
\label{sec:results}

This section analyzes the EuraGovExam evaluation results across 28 VLMs
(14 closed-source, 14 open-source).
We examine overall performance and domain-specific patterns (\S\ref{subsec:overall}),
cross-regional performance variance (\S\ref{subsec:regional}),
and the characteristics of questions where all models fail (\S\ref{subsec:universal_failure}).

\subsection{Overall Performance}
\label{subsec:overall}

Table~\ref{tab:main_results} reports the performance of all evaluated models
under a unified image-only protocol (\S\ref{sec:protocol}).

\paragraph{Closed-source models.}
The top three closed-source models---Gemini-2.5-Pro (86.99\%), GPT-5 (85.80\%), and o3 (84.26\%)---exceed 80\% accuracy across most domains, with Gemini-2.5-Pro surpassing 80\% in 15 of 17 tasks.
However, substantial variability persists within model families:
GPT-4o attains only 42.04\%, trailing GPT-4.1 (54.73\%) by 12.7 percentage points,
highlighting the role of multilingual and non-Latin-script document coverage in image-only reasoning.

\paragraph{Open-source models.}
Open-source models lag considerably: the strongest, Qwen2-VL-72B (44.65\%), trails Gemini-2.5-Pro by over 42 percentage points.
Several widely used models fall below the random baseline (23.7\%), including LLaMA-3.2-11B-Vision (12.75\%) and LLaVA-NeXT-Video-7B (7.21\%), exhibiting near-uniform accuracy across nations and domains indicative of stochastic guessing.
This gap---substantially larger than those on text-only benchmarks such as MMLU~\cite{hendrycks2021measuring}---suggests that image-only multilingual evaluation disproportionately penalizes models with weaker visual encoders and limited script coverage.

\paragraph{Domain-specific patterns.}
Domain difficulty rankings remain remarkably consistent across models.
\textit{High-performing domains} include Philosophy, Psychology, and Language,
where top models frequently exceed 90\% accuracy.
In contrast, \textit{low-performing domains}---notably Earth Science, Geography, and Law---exhibit
systematic degradation, characterized by dense tables, charts, symbolic notations, and spatial references
that require precise visual parsing and cross-modal alignment.

\subsection{Cross-Regional Performance Variance}
\label{subsec:regional}

The central finding of this study is that
\textbf{nation/region induces substantially greater performance variance than domain}.
This phenomenon is consistently observed through within-model cross-regional gaps,
model$\times$region interaction patterns,
and variance decomposition analysis.

\begin{figure*}[t]
\centering
\includegraphics[width=\textwidth]{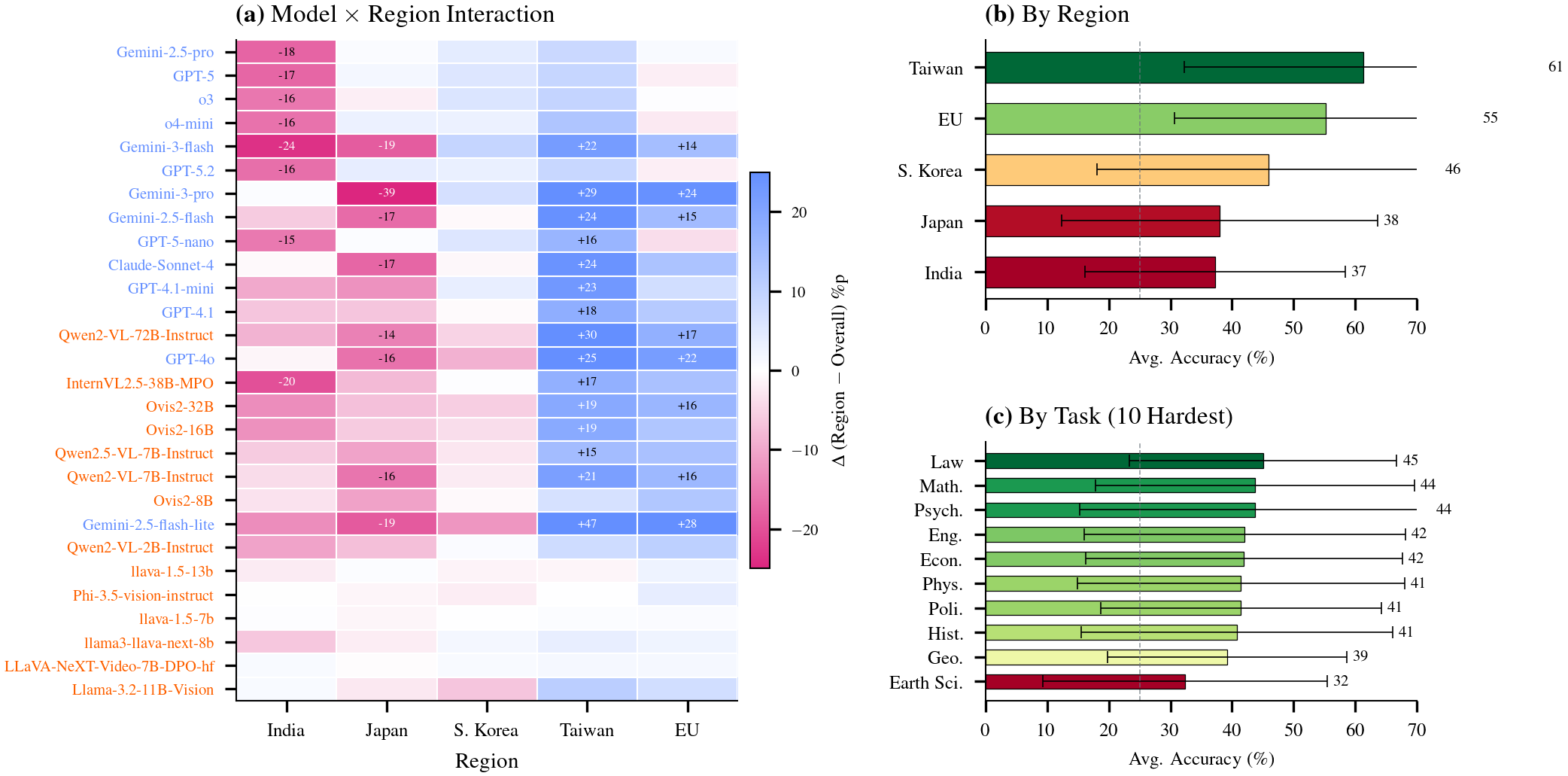}
\caption{\textbf{Cross-regional performance analysis.}
(a) Model$\times$Region interaction heatmap: each cell shows $\Delta = \text{Region} - \text{Overall}$ accuracy;
magenta = underperformance, blue = overperformance (values shown for $|\Delta| \geq 14$\%p).
Model names are color-coded by source type (blue: closed, orange: open).
(b) Average accuracy by region with 95\% Wilson CIs.
(c) Ten most difficult domains.
}
\label{fig:regional_analysis}
\end{figure*}

\paragraph{Regional performance distribution.}
Figure~\ref{fig:regional_analysis}(b) presents the model-averaged accuracy across five regions.
A consistent regional ranking emerges across all 28 models:
Taiwan (57\%) ranks highest,
followed by EU (52\%), South Korea (42\%),
India (38\%), and Japan (35\%).
Table~\ref{tab:nation_accuracy} presents nation-specific accuracy for six representative models
with Wilson 95\% confidence intervals.

\paragraph{Extreme cross-regional gaps.}
Within-model performance gaps across nations are remarkably large.
For Gemini-3-Pro, we observe a \textbf{64.8 percentage point} difference
between Taiwan (96.1\%) and Japan (31.3\%),
indicating that the same model can exhibit over 3$\times$ performance variation
depending on the country of origin.
This pattern is not limited to a single model:

\begin{table}[ht]
\centering
\caption{Nation-specific accuracy (\%) for selected models.
$\pm$: Wilson 95\% CI.
\textbf{Bold}: highest region;
\underline{underline}: lowest region.
Gemini-3-Pro$^\dagger$: largest cross-regional gap (64.8\%p).}
\label{tab:nation_accuracy}
\vspace{0.3em}
\renewcommand{\arraystretch}{1.1}
\resizebox{0.47\textwidth}{!}{%
\begin{tabular}{@{}l c ccccc@{}}
\toprule
\textbf{Model} & \textbf{Ovr.} & \textbf{JP} & \textbf{IN} & \textbf{KR} & \textbf{EU} & \textbf{TW} \\
\midrule
Gemini-2.5-Pro & 87.0 & 87.6\tiny{$\pm$1.3} & \underline{69.2}\tiny{$\pm$2.2} & 91.1\tiny{$\pm$1.3} & 88.1\tiny{$\pm$1.8} & \textbf{95.5}\tiny{$\pm$1.7} \\
GPT-5          & 85.8 & 87.5\tiny{$\pm$1.3} & \underline{68.4}\tiny{$\pm$2.2} & 91.2\tiny{$\pm$1.3} & 83.9\tiny{$\pm$2.0} & \textbf{94.8}\tiny{$\pm$1.9} \\
o3             & 84.3 & 82.4\tiny{$\pm$1.5} & \underline{68.6}\tiny{$\pm$2.2} & 90.1\tiny{$\pm$1.4} & 84.5\tiny{$\pm$2.0} & \textbf{93.7}\tiny{$\pm$2.0} \\
o4-mini        & 79.4 & 82.5\tiny{$\pm$1.5} & \underline{63.4}\tiny{$\pm$2.3} & 82.5\tiny{$\pm$1.7} & 77.0\tiny{$\pm$2.3} & \textbf{92.3}\tiny{$\pm$2.2} \\
Gem.-3-flash   & 75.3 & 56.7\tiny{$\pm$1.9} & \underline{51.7}\tiny{$\pm$2.4} & 84.5\tiny{$\pm$1.6} & 89.5\tiny{$\pm$1.7} & \textbf{97.3}\tiny{$\pm$1.4} \\
\midrule
Gem.-3-pro$^\dagger$ & 68.5 & \underline{31.3}\tiny{$\pm$2.0} & 66.5\tiny{$\pm$2.9} & 74.4\tiny{$\pm$1.7} & 89.0\tiny{$\pm$1.4} & \textbf{96.1}\tiny{$\pm$1.6} \\
\bottomrule
\multicolumn{7}{l}{\tiny $^\dagger$Largest gap: TW--JP = 64.8\%p.}
\end{tabular}%
}
\end{table}

Gemini-2.5-Flash-Lite (66.2\%p),
Gemini-3-Flash (45.6\%p), and GPT-4o (40.7\%p)
all show gaps exceeding 40 percentage points.
Crucially, despite both Japan and Taiwan belonging to the Chinese character (Hanzi/Kanji) cultural sphere,
Taiwan consistently yields higher performance across all models.
This phenomenon cannot be explained by simple linguistic differences alone,
suggesting the influence of \textbf{script-level characteristics}---Japanese vertical writing,
mixed Kanji-Kana text, and furigana annotations---as well as
\textbf{document layout} factors.

\paragraph{Model$\times$region interaction.}
The heatmap in Figure~\ref{fig:regional_analysis}(a) visualizes
the relative strengths and weaknesses of individual models across regions.
Most models exhibit negative $\Delta$ (relative underperformance) in India and Japan,
while showing positive $\Delta$ (relative overperformance) in Taiwan and EU.
Gemini-3-Pro displays the most extreme imbalance,
recording $-39$\%p in Japan and $+29$\%p in Taiwan.
Notably, the two newest Google models (Gemini-3-Pro and Gemini-3-Flash) both exhibit
severe Japan underperformance ($-39$\%p and $-19$\%p respectively),
suggesting that Japanese document processing remains a persistent challenge
that has not been resolved in newer model generations.

\paragraph{Variance decomposition: Nation vs.\ Domain.}
To quantify the sources of VLM performance variation,
we compute nation-wise variance ($\sigma^2_{\text{nation},m}$) and
domain-wise variance ($\sigma^2_{\text{task},m}$) for each model $m$.
Computing the arithmetic mean of variances across all 28 models,
we find that nation factors induce \textbf{2.50$\times$} greater variance than domain factors:
\begin{equation}
\frac{\bar{\sigma}^2_{\text{nation}}}{\bar{\sigma}^2_{\text{task}}}
= \frac{160.53}{64.20} = 2.50
\label{eq:variance_ratio}
\end{equation}
This ratio is even more pronounced for lower-performing models:
for Gemini-2.5-Flash-Lite,
nation variance reaches \textbf{5.5$\times$} the domain variance.

\subsection{Universal Failure Analysis}
\label{subsec:universal_failure}

\begin{table}[ht]
\centering
\caption{Universal failure questions (all 28 models incorrect). India accounts for 60\% despite 12.7\% of the dataset.}
\label{tab:universal_failure}
\vspace{0.3em}
\renewcommand{\arraystretch}{1.05}
\resizebox{0.75\columnwidth}{!}{
\begin{tabular}{@{}lcc|lc@{}}
\toprule
\textbf{Nation} & \textbf{Cnt} & \textbf{\%} & \textbf{Domain} & \textbf{Cnt} \\
\midrule
India    & 30 & 2.9 & Mathematics    & 9 \\
S. Korea &  8 & 0.3 & Economics      & 7 \\
Japan    &  6 & 0.3 & Engineering    & 6 \\
EU       &  6 & 0.3 & Administration & 5 \\
Taiwan   &  0 & 0.0 & History        & 4 \\
\midrule
\textbf{Total} & \textbf{50} & \textbf{0.6} & Other (11)  & 15 \\
\bottomrule
\end{tabular}%
}
\end{table}

Out of 8,000 questions, 50 (0.6\%) elicit incorrect responses from every evaluated model (Table~\ref{tab:universal_failure}).
India accounts for 60\% of these failures (30/50) despite comprising only 12.7\% of the dataset (failure rate 2.9\% vs.\ 0.3\% elsewhere), while Taiwan contributes zero.
Qualitative examination reveals three co-occurring characteristics: complex script environments (non-Latin characters within mathematical expressions), dense visual elements (nested tables, small-font annotations), and domain-specific symbols (circuit diagrams, chemical formulas) demanding joint visual and domain knowledge.

Notably, universal failures concentrate in India rather than Japan, which exhibits the lowest \textit{average} accuracy---revealing a distinction between \textit{difficulty} and \textit{intractability}.
Japanese questions are systematically harder but partially solved by top models; Indian failures defeat even the strongest systems, suggesting a deeper entanglement of perception, domain knowledge, and culturally specific reasoning.

\section{Discussion}
\label{sec:discussion}

\paragraph{Why region dominates domain.}
The 2.50$\times$ variance ratio (Eq.~\ref{eq:variance_ratio}) carries a direct implication:
\textbf{benchmarks that control for domain but not for script and layout may substantially underestimate real-world performance gaps.}
A model scoring 90\% on Philosophy from Taiwan and the EU can fall below 30\% on the same domain from Japan---not because domain knowledge differs, but because the visual presentation does.
Our qualitative failure analysis (Appendix~B) provides the causal layer behind this pattern: Japan's errors concentrate in visual parsing failures driven by vertical text layouts and dense furigana annotations, whereas India's errors split between reasoning and knowledge deficits tied to Devanagari-embedded technical content.
These region-specific bottleneck profiles confirm that a single remediation strategy will yield uneven returns across regions, and that the diagnostic granularity enabled by \textsc{EuraGovExam}'s factorial structure is unavailable from aggregate-accuracy benchmarks.

\paragraph{Directions for multilingual VLM development.}
Our results suggest three concrete improvements.
First, \textit{script-aware visual encoding}: the persistent Japan--Taiwan gap despite shared Hanzi/Kanji origins shows that character-level similarity does not guarantee comparable processing; Japanese vertical text, mixed Kanji-Kana layouts, and furigana annotations demand layout-aware tokenization strategies.
Second, \textit{balanced multilingual training data}: the correlation between regional performance and likely training-data availability suggests that targeted augmentation for underrepresented scripts---particularly Japanese document scans and Devanagari-embedded technical content---could yield disproportionate gains, especially since newer model generations have not closed these gaps through scaling alone.
Third, \textit{document layout as a first-class evaluation axis}: the consistent underperformance of layout-dense regions indicates that layout complexity should be varied independently alongside domain and language in future benchmarks.

\paragraph{Limitations.}
We identify four principal limitations.
\emph{First}, the benchmark covers five regions spanning diverse script families and governance traditions but does not include Africa, the Middle East, or Latin America; findings about script-driven bottlenecks may not generalize to Arabic or Amharic writing systems.
\emph{Second}, regional sources differ in domain composition and question difficulty, reflecting genuine differences in national examination design rather than controlled experimental variation.
While our two-way ANOVA partially accounts for this confound, strictly causal claims about script versus domain effects require further controlled experiments.
\emph{Third}, the image-only protocol precludes disentangling OCR-level failures from higher-level visual understanding errors; a model that extracts text perfectly but misinterprets a diagram and one that fails at character recognition both manifest as incorrect answers.
\emph{Fourth}, our benchmark captures a static snapshot of government examinations; regulatory and curricular changes may alter difficulty distributions over time, necessitating periodic updates to maintain ecological validity.

\paragraph{Future directions.}
Three research directions emerge from this work.
\emph{Intervention-based diagnosis:}
systematically varying input modality---providing OCR-extracted text alongside images, or translating questions into English---would decompose performance gaps into perception, language, and reasoning components, transforming \textsc{EuraGovExam} from a diagnostic benchmark into an intervention-testing platform.
Our image-only protocol establishes the baseline against which such interventions can be measured.
\emph{Geographic and temporal expansion:}
extending to additional script families (Arabic, Thai, Cyrillic) and tracking examination changes over time would enable longitudinal studies of whether geographic biases are closing or widening as models scale.
\emph{Fine-grained annotation:}
augmenting each question with visual complexity scores, required knowledge type, and reasoning depth would support targeted data augmentation for underperforming region--domain cells.
We release our evaluation toolkit and annotation guidelines to facilitate community contributions along these axes.

\section{Conclusion}
\label{sec:conclusion}

We presented \textsc{EuraGovExam}, an image-only benchmark
of over 8,000 government-examination questions drawn from
five Eurasian regions (South Korea, Japan, Taiwan, India, and the EU)
across 17 professional domains and five distinct script families.
Unlike existing multilingual benchmarks that supply OCR-extracted
text or translated questions, our protocol preserves original
document layouts and scripts, requiring models to jointly perceive
and reason over authentic visual content.

Evaluation of 28 VLMs yields two principal findings.
\emph{First}, nation and script factors induce
2.50$\times$ greater performance variance than domain factors,
with cross-regional gaps reaching 64.8 percentage points
within a single model---demonstrating that the visual and
linguistic properties of documents, not just their content,
are a primary determinant of VLM performance.
\emph{Second}, 50 questions defeat all 28 models,
concentrating in India (60\%)
and co-occurring in visually complex, knowledge-intensive items
that resist improvement through scaling alone.

These results carry a practical implication:
evaluation protocols that control for domain but not for
script and layout will systematically underestimate
real-world performance gaps.
\textsc{EuraGovExam}'s factorial nation$\times$domain structure
provides the diagnostic resolution needed to identify
\emph{where} and \emph{why} models fail,
guiding targeted improvements in script-aware encoding,
multilingual training balance, and layout-robust architectures.

We release the full dataset, evaluation scripts, and model outputs
to support further research.\footnote{%
Dataset: \url{https://huggingface.co/datasets/EuraGovExam/EuraGovExam};
  
Code: 
\url{https://github.com/thisiskorea/EuraGovExam};
  
Website:
\url{https://thisiskorea.github.io/EuraGovExam/};}

\begin{acks}
This work was supported by the Ministry of Science and ICT and the
National Research Foundation of Korea (NRF) grant funded by the
Korean government (MSIT) (No.~RS-2026-25498341).
\end{acks}

\bibliographystyle{ACM-Reference-Format}
\makeatletter
\def\balance{}
\makeatother
\bibliography{sample-base}

\clearpage

\appendix

\section{Additional Analyses: Scaling, Task Difficulty, and Cross-Task Structure}
\label{app:analysis}

\paragraph{Overview.}
This appendix complements the main results (overall accuracy and country/domain variability) with a set of \emph{diagnostic analyses}.
Specifically, we quantify (i) \textbf{scaling trends} among open-source multimodal models, (ii) \textbf{task-level difficulty structure} across model performance tiers, and (iii) \textbf{cross-task covariance and latent structure} via correlation-based clustering.
Together, these analyses support the view that our benchmark measures not only a single aggregate score but also \emph{structured difficulty patterns} and \emph{latent capability axes}.

\subsection{Scaling Trends in Open-Source Models}
\label{app:analysis:scaling}

\begin{figure}[ht]
  \centering
  \includegraphics[width=0.95\linewidth]{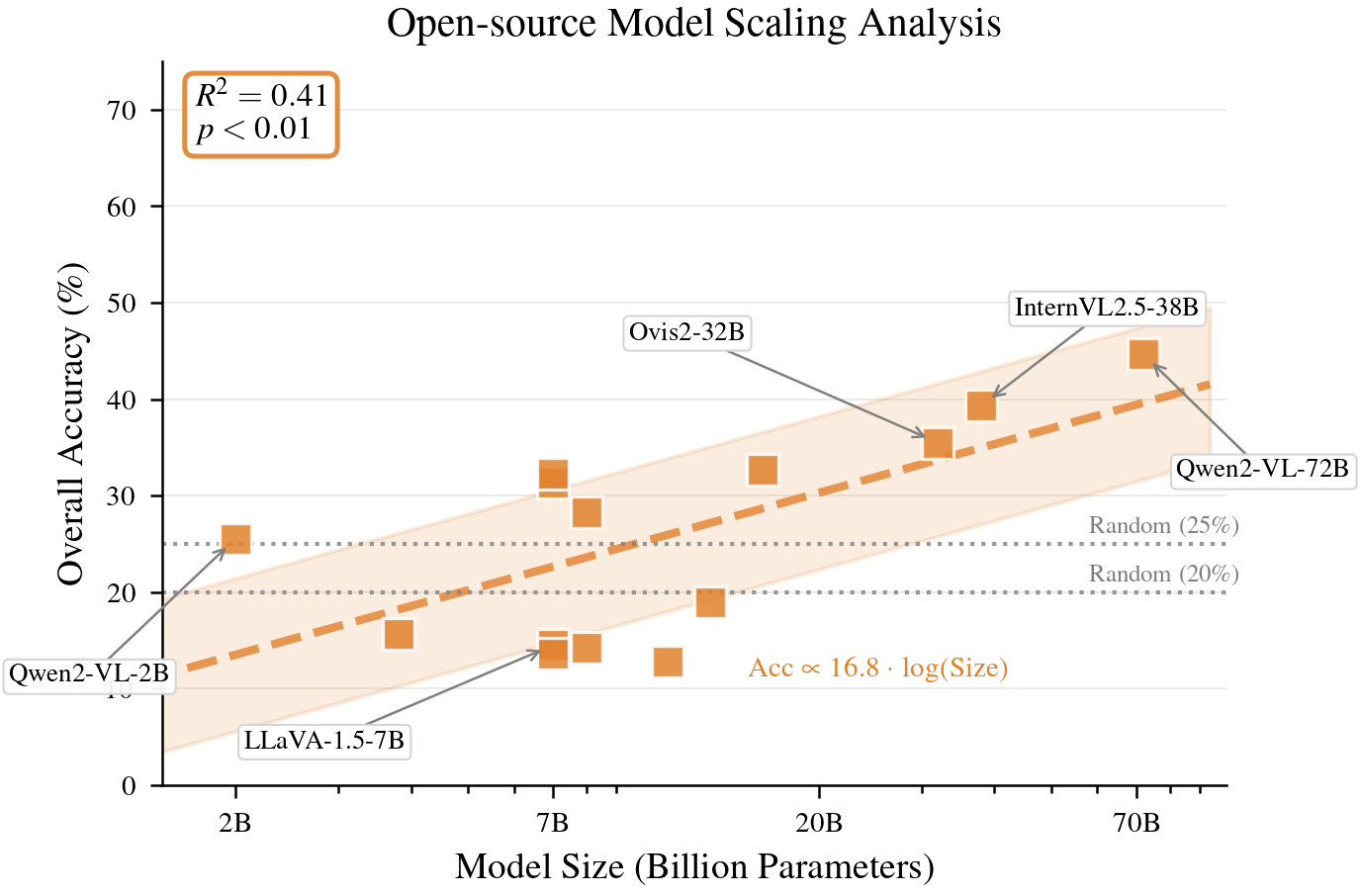}
  \caption{\textbf{Scaling analysis for open-source models.}
  We analyze the relationship between parameter scale (in billions, $S$) and overall benchmark accuracy ($A$) for open-source VLMs using a linear regression in log scale.
  The dashed line shows the fitted trend, and the shaded region indicates a $\pm 1$ standard-deviation band of the regression residuals (an empirical $\pm 1$ SD band).
  We obtain $R^2=0.41$ with $p<0.01$, and observe an empirical trend of the form $A \approx 16.8\cdot \log(S)$.}
  \label{fig:app:scaling}
\end{figure}

\paragraph{Setup.}
For the open-source model set $\mathcal{M}_{\text{open}}$, let $S_m$ denote the parameter count of model $m$ (in billions) and $A_m$ its overall benchmark accuracy.
To characterize scaling behavior, we fit the following linear model:
\begin{equation}
A_m = \beta_0 + \beta_1 \log(S_m) + \epsilon_m,
\label{eq:app:scaling}
\end{equation}
where $\epsilon_m$ is the residual. We test the regression significance using a standard $t$-test under the usual assumptions.

\paragraph{Observation.}
Figure~\ref{fig:app:scaling} shows that accuracy increases on average with $\log(S)$, but the explanatory power remains moderate ($R^2=0.41$).
This indicates that \textbf{parameter scale alone does not account for the full variance in performance}, and that models with comparable size can still differ substantially.

\paragraph{Implication.}
Beyond raw scale, our benchmark is sensitive to differences in factors such as
\emph{document visual processing (layout/text), training recipe, and multimodal encoder design}.
In other words, while scaling contributes to improved aggregate scores, \textbf{practical bottlenecks may arise from non-scale capability axes}
(e.g., document understanding, quantitative reasoning, or knowledge integration).

\subsection{Task Difficulty Structure Across Model Tiers}
\label{app:analysis:tier}

\begin{figure}[ht]
  \centering
  \includegraphics[width=0.95\linewidth]{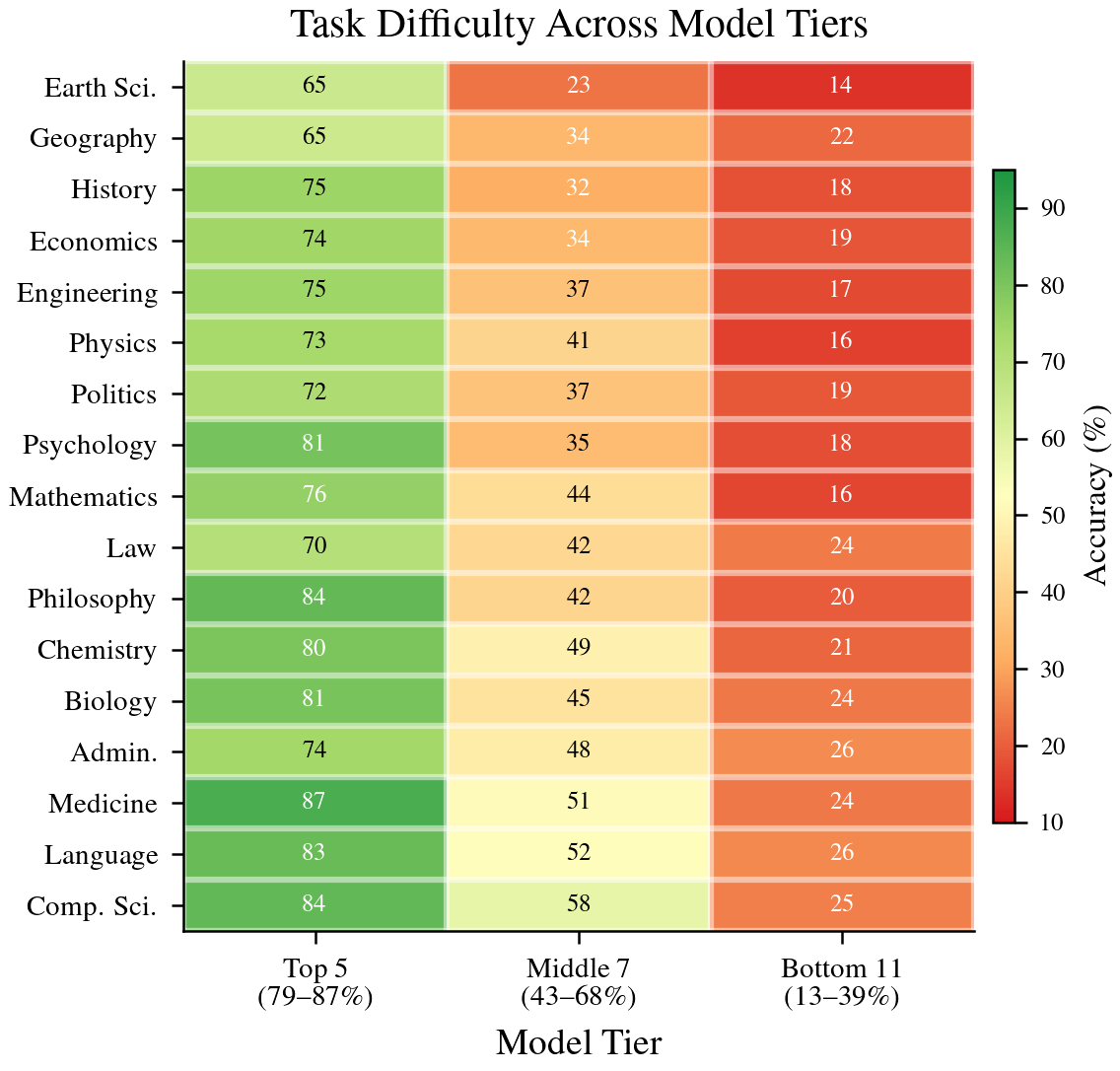}
  \caption{\textbf{Task difficulty by model tier (mean accuracy).}
  We partition models into Top, Middle, and Bottom tiers by overall accuracy, and visualize per-task mean accuracy as a heatmap.
  This reveals whether the relative ordering of task difficulty is preserved as models improve, and which tasks remain persistent bottlenecks.}
  \label{fig:app:tier}
\end{figure}

\begin{figure*}[!t]
  \centering
  \includegraphics[width=0.85\linewidth]{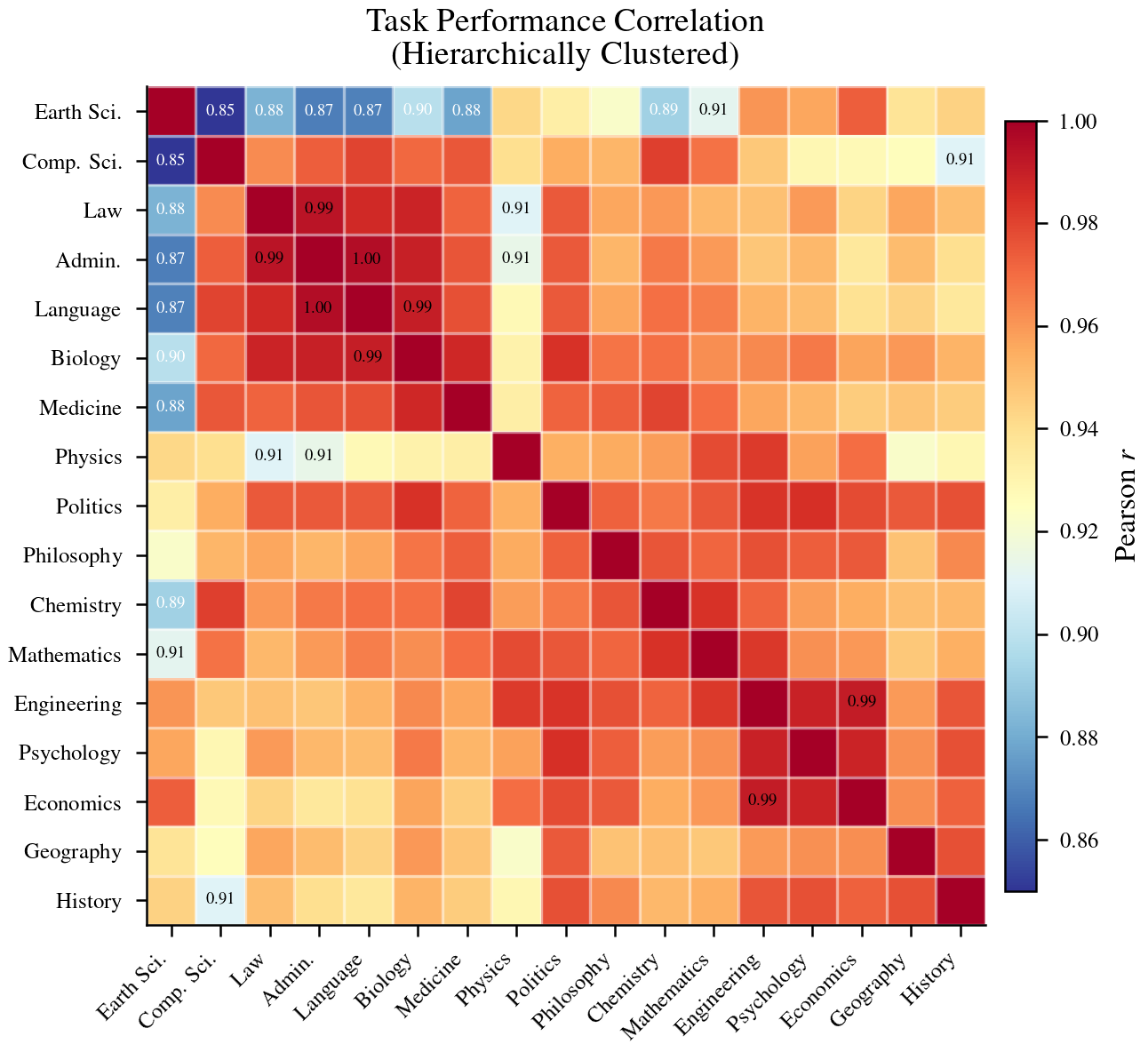}
  \caption{\textbf{Cross-task correlation with hierarchical clustering.}
  For each task, we form a \emph{model-wise accuracy vector} and compute Pearson correlations $r(t,t')$ between tasks.
  We then apply hierarchical clustering (average linkage) using Euclidean distance between \emph{correlation profiles} (rows/columns of the correlation matrix).
  High-correlation blocks suggest that subsets of tasks co-vary due to shared \emph{latent capability axes}.}
  \label{fig:app:corr}
\end{figure*}

\paragraph{Tier definition.}
We sort models in $\mathcal{M}$ by overall accuracy $\mathrm{Acc}(m)$ and construct three tiers:
the top $K_{\text{top}}$ models $\mathcal{M}_{\text{top}}$,
the middle $K_{\text{mid}}$ models $\mathcal{M}_{\text{mid}}$,
and the bottom $K_{\text{bot}}$ models $\mathcal{M}_{\text{bot}}$.
In our experiments, we use \emph{Top 5 / Middle 7 / Bottom 11}.

For each task $t \in \mathcal{T}$ and tier $\tau \in \{\text{top, mid, bot}\}$, we define the tier-averaged task accuracy as
\begin{equation}
\overline{A}(t,\tau)
= \frac{1}{|\mathcal{M}_\tau|}\sum_{m\in\mathcal{M}_\tau} A(m,t),
\label{eq:app:tier}
\end{equation}
where $A(m,t)$ is the accuracy of model $m$ on task $t$.

\paragraph{Observation.}
Figure~\ref{fig:app:tier} highlights (i) tasks where performance collapses sharply in the Bottom tier,
(ii) \emph{hard-core} tasks that remain relatively low even in the Top tier,
and (iii) groups of tasks whose relative difficulty ordering is largely preserved across tiers.
Notably, steep drops in the Bottom tier are most pronounced for tasks that \textbf{require both document-level visual understanding (layout/text) and domain-specific reasoning}.

\paragraph{Implication.}
Tier-wise heatmaps expose \textbf{bottleneck tasks that are invisible from the mean score alone}.
Moreover, if difficulty structure remains stable across tiers, this suggests our benchmark is not a noisy collection of items but rather provides
\textbf{a consistent, structurally decomposable evaluation signal}.

\subsection{Cross-Task Correlation and Latent Structure (Clustering)}
\label{app:analysis:corr}

\paragraph{Correlation matrix.}
For each task $t$, let the model-wise accuracy vector be
$\mathbf{a}_t = \big(A(m_1,t), \dots, A(m_{|\mathcal{M}|},t)\big) \in \mathbb{R}^{|\mathcal{M}|}$.
We define the Pearson correlation between tasks $t$ and $t'$ as
\begin{equation}
r(t,t') = \mathrm{corr}\big(\mathbf{a}_t, \mathbf{a}_{t'}\big).
\label{eq:app:corr}
\end{equation}
Let $\mathbf{R}=[r(t,t')]_{t,t'\in\mathcal{T}}$ be the resulting correlation matrix.
To cluster tasks, we compute Euclidean distance between correlation profiles (rows of $\mathbf{R}$):
\begin{equation}
d_{\text{row}}(t,t') = \left\lVert \mathbf{R}_{t,:} - \mathbf{R}_{t',:}\right\rVert_2,
\label{eq:app:corr_rowdist}
\end{equation}
and perform average-linkage hierarchical clustering to determine the dendrogram ordering.
(Importantly, our ordering is \emph{not} based on $d=1-r$, but on similarity of full correlation profiles.)

\paragraph{Observation.}
The block structure in Figure~\ref{fig:app:corr} indicates that tasks do not vary independently across models:
instead, \textbf{some task groups rise and fall together}.
This suggests that overall performance may be explained by a combination of
\textbf{global document-understanding capability} plus a small number of \textbf{specialized axes}
(e.g., quantitative reasoning, regulation/linguistic inference).

\paragraph{Implication.}
The correlation and clustering structure strengthens the benchmark's \textbf{diagnostic value}.
(1) Beyond the main aggregate score, we can report vulnerability patterns at the cluster level, enabling more precise characterization of model bottlenecks.
(2) It supports targeting data collection and model improvements at the level of \emph{latent capability axes}, rather than treating tasks as isolated endpoints.
Finally, broadly high correlations would be consistent with the benchmark measuring \textbf{a coherent set of document-grounded competencies}.

\section{Qualitative Failure Analysis}
\label{app:failure_cases}

To examine \emph{how} and \emph{why} state-of-the-art VLMs fail on
EuraGovExam, we sampled ten representative cases---two per region---and
classified each along the four error axes used throughout the paper:
\textbf{Visual Parsing (VP)}, \textbf{Reasoning (RF)}, \textbf{Knowledge
(KF)}, and \textbf{Hallucination (HC)}.
Table~\ref{tab:case-summary} reports the resulting taxonomy together
with the predicted and ground-truth answer choices.
 
\begin{table}[ht]
\centering
\caption{Summary of qualitative failure cases.
\textbf{VP}: Visual Parsing, \textbf{RF}: Reasoning,
\textbf{KF}: Knowledge, \textbf{HC}: Hallucination.
``--'' in the \textbf{Pred} column indicates that the model failed to
produce a valid answer choice.}
\label{tab:case-summary}
\setlength{\tabcolsep}{3pt}
\renewcommand{\arraystretch}{1.05}
\resizebox{\columnwidth}{!}{%
\begin{tabular}{@{}c l l l l c@{}}
\toprule
\textbf{\#} & \textbf{Region} & \textbf{Domain} & \textbf{Type} & \textbf{Model} & \textbf{Pred (GT)} \\
\midrule
1  & S.~Korea & Economics      & KF          & Gemini-2.5-Pro        & D (A) \\
2  & S.~Korea & Administration & RF          & Claude-Sonnet-4       & B (D) \\
3  & Japan    & Physics        & RF          & o4-mini               & B (C) \\
4  & Japan    & Engineering    & KF\,+\,VP   & Qwen2-VL-72B          & B (C) \\
5  & India    & Mathematics    & RF\,+\,HC   & Llama-3.2-11B-Vision  & A (B) \\
6  & India    & Economics      & VP          & LLaVA-1.5-13B         & -- (C) \\
7  & EU       & Administration & KF          & Claude-Sonnet-4       & -- (D) \\
8  & EU       & Administration & VP\,+\,HC   & LLaMA3-LLaVA-NeXT-8B  & -- (A) \\
9  & Taiwan   & Administration & RF          & Phi-3.5-Vision        & -- (A) \\
10 & Taiwan   & Biology        & KF          & GPT-4.1               & B (D) \\
\bottomrule
\end{tabular}%
}
\end{table}
 
\paragraph{Script and layout dominate over domain reasoning.}
Six of the ten cases (\#3,~4,~5,~6,~8,~9) hinge on perception of the
\emph{source script or layout}---vertical Japanese typesetting,
Devanagari headers, Traditional Chinese statutory text, or multi-panel
figure composition---rather than on domain content. Only four cases
(\#1,~2,~7,~10) are domain-specific knowledge or reasoning gaps that
surface even when perception is intact. This 6:4 split closely mirrors
the $2.52\times$ nation/script-over-domain variance reported in the
main results, and shows that the same effect persists at the
individual-question level rather than averaging out across questions.
 
\paragraph{Failures concentrate at the perception--reasoning interface.}
Pure single-axis failures (only VP, only RF, only KF) account for six
of the ten cases; the remaining four (\#4 KF\,+\,VP, \#5 RF\,+\,HC,
\#8 VP\,+\,HC, and the answer-omissions in \#6--9) involve two or more
co-occurring errors. In every multi-axis case, a perceptual mistake
upstream---a misread axis, a hallucinated constraint, a misparsed
footnote---propagates into a downstream reasoning or knowledge step
that, in isolation, the same model handles correctly elsewhere in the
benchmark. This cascading pattern is consistent across both the
strongest closed-source models (Claude-Sonnet-4, Gemini-2.5-Pro,
GPT-4.1, o4-mini) and the weaker open-source models, suggesting that
robustness at the perception--reasoning boundary, rather than raw
reasoning capacity, is the binding constraint on EuraGovExam
performance.
 
\paragraph{No-valid-answer failures expose pipeline brittleness.}
Four cases (\#6, \#7, \#8, \#9) yield no valid answer choice at all:
the model either omits a final letter or invents an option that does
not appear in the stem. These breakdowns cluster in EU and Taiwan
administration items and in Indian economics, all of which combine
non-Latin script with dense, regulation-style multi-column layouts.
The fact that strong models such as Claude-Sonnet-4 fall into this
category alongside small open-source models indicates that the
failure is not simply a parameter-scale phenomenon but a property of
the input distribution: when layout complexity and script unfamiliarity
co-occur, the end-to-end pipeline can collapse rather than degrade
gracefully.
 
\paragraph{Implications for benchmark design.}
The qualitative pattern reinforces three design choices in
EuraGovExam. \emph{First}, embedding all question content within a
single image is what surfaces the perception--reasoning cascades; a
text-extracted variant would silently bypass cases \#4, \#6, and \#8.
\emph{Second}, preserving the original script and layout exposes
script-specific bottlenecks that are invisible in translated or
romanized benchmarks. \emph{Third}, retaining authentic
jurisdiction-specific administrative content (KDC classes, EU treaty
articles, Korean public-sector accounting) makes the residual KF/RF
failures diagnostically meaningful rather than artefacts of synthetic
question construction. Together, these cases support the main paper's
recommendation that next-generation VLMs be evaluated under
script-faithful, layout-faithful, image-only conditions.

\end{document}